\documentclass[10pt]{article} 

\usepackage[accepted]{rlj} 

\usepackage{amssymb}            
\usepackage{mathtools}          
\usepackage{mathrsfs}           
\usepackage{graphicx}           
\usepackage{subcaption}         
\usepackage[space]{grffile}     
\usepackage{url}                
\usepackage{lipsum}             

\usepackage{amsmath}
\DeclareMathOperator*{\argmax}{arg\,max}

\usepackage{mathrsfs}           
\usepackage{graphicx}           
\usepackage{subcaption}         

\usepackage{comment}
\usepackage{booktabs}
\usepackage{float}

\newcommand{\MET}[1]{\textcolor{blue}{MET: #1}}

\newcommand{\svec}{\mathbf{s}}
\newcommand{\avec}{\mathbf{a}}
\newcommand{\cvec}{\mathbf{c}}

\newcommand{\hvec}{\mathbf{h}}
\newcommand{\ovec}{\mathbf{o}}
\newcommand{\wvec}{\mathbf{w}}

\title{Efficient Morphology-Aware Policy Transfer to New Embodiments}

\setrunningtitle{Efficient Morphology-Aware Policy Transfer to New Embodiments}

\author{Michael Przystupa$^{1,3,4}$, Hongyao Tang$^{3,4}$, Martin Jagersand$^{1}$, Santiago Miret$^{5}$, Mariano Phielipp$^{5}$, Matthew E.~Taylor$^{1,2}$, Glen Berseth$^{3,4}$ }

\emails{przystup@ualberta.ca}

\affiliations{
$^{1}$\textbf{ University of Alberta, Canada},
$^{2}$\textbf{Alberta Machine Intelligence Institute (Amii)}, 
$^{3}$\textbf{Université de Montréal}, 
$^{4}$ \textbf{Mila Institute}, 
$^{5}$ \textbf{Intel Labs}
\par 
}

\contribution{
    We conduct an extensive series of experiments to compare the effects of parameter-efficient finetuning methods in the morphology-aware policy learning setting. 
    }
    {
    Prior works which include transfer learning experiments have generally focused on end-to-end finetuning or else at most consider low-rank adapter layers (LoRA), a form of delta weight learning, as part of their experiments \citep{octoteam2024OctoOpensourcegeneralist}. When LoRA has been used, experiments have only been conducted only in the \textit{behavioral cloning} setting. This is a limitation in the literature because a wide variety of parameter-efficient techniques have been investigated in other fields such as prefix tuning in large language models \citep{LiL20Prefix-tuning} and direct-finetuning in computer vision \citep{lee2022surgical}.
    
    }

\contribution{
    We are the first work to successfully learn policies using prefix tuning methods 
    in the reinforcement learning settings. 
    }
    {
    Prefix tuning has been almost exclusively investigate in supervised learning settings such as natural language processing \citep{LiL20Prefix-tuning}, computer vision \citep{nie2023protuningUnifiedvision}, or continual learning \citep{Wang2022promptingInContinualLearning}. The closest related to our work is 
    \cite{LiuZA0ZSF24TAIL} who investigate prefix tuning techniques in the imitation learning setting and across \textit{tasks} as opposed to agent morphology. 
    }

\contribution{
Our experiments reveal a number of trends in the morphology-aware policy setting. Generally we find that both input-adapter and prefix tuning methods converge to behaving similar to tuning the decoder head of the base model. Prefix tuning is particularly sensitive to hyperparameter choices where some configurations notably affect performance at the beginning of training and never recover. Generally, more parameters are always beneficial to improving policy performance in the tasks we considered. 
}{
Other such prescriptive research has been done in computer vision or language when investigating different PEFT techniques.  The work of \cite{lester2021powerscaleforPEFT} demonstrated the potential of prefix tuning over a number of factors including prompt initialization and number of prompt tokens. The work of \cite{liu2022ptuningprmopting} highlights the benefits of injecting prompts in multiple layers in transformers. The work of \cite{lee2022surgical} suggests that intelligent layer different types of domain shifts in computer vision.

}


\summary{

In this work, we investigate means of reducing the computation costs to finetune pre-trained morphology-aware policies to target morphologies with on-policy learning. Morphology-aware learning is a paradigm which attempts to learn several optimal policies across agent embodiments in a \textit{single} neural network. A limitation of prior works have been focusing on end-to-end finetuning to adapt these policies to a target morphology. We address this gap by exploring parameter efficient techniques used successfully in other domains such as computer vision or natural language processing to specialize a policy. Our results suggest that using as few as 1\% of total learnable parameters as the pre-trained model, we can achieve statistically significant performance improvements. 

}

\begin{document}

\makeCover  
\maketitle  

\begin{abstract}

Morphology-aware policy learning is a means of enhancing policy sample efficiency by aggregating data from multiple agents. These types of policies have previously been shown to help generalize over dynamic, kinematic, and limb configuration variations between agent morphologies. Unfortunately, these policies still have sub-optimal zero-shot performance compared to end-to-end finetuning on morphologies at deployment. This limitation has ramifications in practical applications such as robotics because further data collection to perform end-to-end finetuning can be computationally expensive. In this work, we investigate combining morphology-aware pretraining with \textit{parameter efficient finetuning} (PEFT) techniques to help reduce the learnable parameters necessary to specialize a morphology-aware policy to a target embodiment. We compare directly tuning sub-sets of model weights, input learnable adapters, and prefix tuning techniques for online finetuning. Our analysis reveals that PEFT techniques in conjunction with policy pre-training generally help reduce the number of samples to necessary to improve a policy compared to training models end-to-end from scratch. We further find that tuning as few as less than 1\% of total parameters will improve policy performance compared the zero-shot performance of the base pretrained a policy. 


\end{abstract}

\section{Introduction}

Learning agents that can reuse knowledge across tasks demonstrate improved sample efficiency and better learning capabilities \citep{reed2022generalist,driess2023palm,deng2023mind2web}. Deep reinforcement learning (RL), despite its potential, faces significant challenges when applied to multiple tasks due to its sensitivity to even minor environmental variations and sample inefficiency \citep{henderson2018deepRLthatMatters,Du2020IsGoodRepresentationSufficient}. Prior research suggests that even subtle dynamic or kinematic differences can notably affect policy performance \citep{ChenMG18Hardware,SchaffYCW19Jointly}. This brittleness and inefficiency create substantial barriers when developing versatile agents that can adapt to new scenarios. Morphology-aware learning is one means of enabling knowledge transfer across different physical agent configurations. Morphology adaptation techniques can improve policy robustness and sample efficiency by explicitly accounting for agent embodiments.


Morphology-aware policy learning incorporates agent morphology knowledge by representing embodiments as graphs processed through GNNs \citep{ScarselliGTHM09GNN} or transformers \citep{VaswaniSPUJGKP17Attention}. Representing agents as graphs is valuable because it enables policies to represent agents with changing limb configurations, and thus varying action spaces \citep{wang2018nervenet,HuangMP20Onepolicy,KurinIRBW21Mybody}.  Research has focused on effective graph structure utilization through adjacency matrices \citep{Hong2022SWAT,LiLZZ24MAT}, feature grouping \citep{trabucco2022anymorph,xiong2023modumorph,sferrazza2024bodytransformer}, and geometric symmetries \citep{chen2023subequivariant}. Morphology-aware learning can improve sample efficiency as supported by theoretical sample bounds in multi-task learning \citep{brunskill2013sample,maurer2016benefit,deramo2020sharing,bohlinger2025onepolicytorun}, with empirical results suggesting policies optimized over morphology distributions outperform specialized ones \citep{gupta2022metamorph,xiong2023modumorph}. Applications include autonomous robot design \citep{PathakLDIE19SelfAssemb,luck2020DataeffCoadptmorphandBehavior,Ye22Transform2Act} and large-scale control models \citep{bousmalis2024robocat,ONeillRMGPLPGMJ24Embodiment-X,octoteam2024OctoOpensourcegeneralist}.

Unfortunately, deploying morphology-aware policies on new embodiments continues to be challenging because of the employment of computationally inefficient transfer learning techniques. Prior works suggest that pre-training morphology-aware policies provide better policy initialization when transferring, but additional finetuning is necessary to elicit optimal performance on new morphologies \citep{gupta2022metamorph,xiong2023modumorph,furuta2023mxtbench}. These works have focused mainly on end-to-end finetuning algorithms, which can be computationally intensive for larger monolithic policies. In resource-constrained settings like robotics \citep{huai2019towardsDLresourceconstraint,neuman2022TinyRobotLearning}, reducing further computation for learning is referable for transferring policies. 

In this work, we investigate parameter-efficient finetuning (PEFT) algorithms as a solution to improve policy transfer performance with reduced computational resources. PEFT algorithms use subsets of a model's parameters to finetune a pre-trained neural network or otherwise introduce a small set of new learning parameters that specialize to a target task \citep{Dong23IncontextSurvey,kirk2023SurveyzeroshotRL}. The latter approach is more flexible because these new parameters can be introduced in ways that do not directly change the pre-trained model \citep{tsai2020transferlearningwithoutknowing}. Researchers have shown that PEFT methods work well on large networks in natural language tasks \citep{LiL20Prefix-tuning} and in computer vision problems \citep{lee2022surgical} while reducing additional computation costs to perform gradient updates on a small set of PEFT parameters compared to the entire model. Closely related to our work is the work of \cite{LiuZA0ZSF24TAIL}, who investigate PEFT methods in continual imitation learning. Our research is different as we deal with \textit{morphology transfer} and evaluate PEFT methods with deep RL, which presents other challenges from supervised learning.

In summary, the primary contribution of our work is the analysis of several PEFT techniques for morphology-aware policy transfer. Our results demonstrate that it is generally achievable to substantially reduce the total parameters used and achieve statistically measurable improvement over zero-shot performance, even with strong initial zero-shot performance. Using even 1\% total learnable parameters relative to the base model's total parameter count leads to measurable performance improvement while significantly reducing learning computation costs compare to end-to-end finetuning. 
As part of our work, we show how input-learnable PEFT algorithms preserve strong zero-shot capabilities as a performance floor and consistently outperform these initial capabilities as training progresses, making them particularly suitable for online reinforcement learning scenarios with limited data collection opportunities. This research has potential in real-world applications like robotic learning. Our results provide practical guidelines for researchers to determine which PEFT techniques best balance sample efficiency, computational requirements, and performance gains for their specific deployment settings.

\section{Background}
\label{sec:background}

\subsection{Contextual Markov Decision Process}

Morphology-aware policy learning can be understood as a form of contextual Markov decision process (CMDP) \citep{hallak2015contextualmarkovdecisionprocesses}. 
A CMDP is characterized by a distribution $\mathcal{C}$, where for $c \sim p(C)$ we have an induced tuple $M(c) = \left( \mathcal{S}^c, \mathcal{A}^c, p^c(s' | s, a), r, p^c(s_0) \right)$. For each $c$, $\mathcal{S}^c$ is a finite set of states, $p^c(s_0)$ represents the initial state distribution, and $\mathcal{A}^c$ is a finite set of actions. The state transition probability function, $p^c(s'| s, a) = \Pr(s_{t+1} = s' \mid s_t = s, a_t = a ; c)$, defines the probability of transitioning from state $s$ to  state $s'$  when action $a$ occurs. The reward function, $r^c(s, a, s')$, represents the immediate value of transitioning from $s$ to $s'$ due to $a$. A policy $\pi: \mathcal{S} \times \mathcal{C} \to \mathcal{P}(\mathcal{A})$ is a mapping from states and contexts to a probability distribution over actions, where $\pi$ samples actions $a \sim \pi(s, c)$ to transition following $p^c(s' | s, a)$. For a given CMDP, the objective is to maximize the expected sum of rewards over the distribution of contexts,
\begin{equation*}
    \pi^\star(s, c) = \argmax_{\pi \in \Pi} \mathbb{E}_{p(c)} [G_c],
\end{equation*}
where $G_c = \mathbb{E}_{p^c(\tau)}[\sum_{t=0}^{T} \gamma^t r(s_t, a_t)]$ is the expected cumulative reward for a given context with discount factor $\gamma \in [0, 1)$. We only consider the finite horizon case where the tasks will terminate after $T \in \mathbb{N}^+$ steps, and $p^c(\tau) = p^c(s_0)\prod_{t=0}^{T} \pi(s_t, c) \mathcal{P}^c(s_{t+1} | s_t, a_t)$ is the distribution over trajectories in the environment.

In our work, the context $c$ is the morphology information, which we represent as a sequence $\cvec \in \mathbb{R}^{l(c) \times d^c}$ which has $l(c) \in \mathbb{N}^+$ limbs and $d^c \in \mathbb{N}^+$ features. Each token $\cvec_{i}$ contains information such as the limb adjacency matrix, link dynamic values (mass, friction, etc.), and link kinematic information (e.g. joint limits and values). We optimize MDPs with continuous action and state spaces, $\avec \in \mathbb{R}^{l(c)}$ and $\svec \in \mathbb{R}^{l(c) \times d^s}$ with $d^s \in \mathbb{N}^+$ are state features. This differs from typical CMDPs which usually assume a fixed dimensionality of states and actions. 


\subsection{Transformers} 

An essential component of the morphology-aware policies in previous works are transformer models \citep{gupta2022metamorph,VaswaniSPUJGKP17Attention,xiong2023modumorph}. We treat our data as an observation sequence $\ovec \in \mathbb{R}^{l(c) \times d}$ with $l(c)$ limb embeddings with $d \in \mathbb{N}^+$ features. Each token $\ovec_i = [\svec_i; \cvec_i]$ contains limb-level state and context variables of a morphology for $i \in [1, 2, ..., l(c)]$.
A morphology-independent linear transformation projects the limb-specific features to a shared embedding space $\bar{\ovec} = \text{LN}(\ovec W^{embed} + W^{position}[1:l])$, where $W^{embed} \in \mathbb{R}^{d \times h}$ is a linear projection operation that transforms the input features to the hidden dimension $h \in \mathbb{N}^+$. $W^{position} \in \mathbb{R}^{L \times h}$ represents the positional embeddings up to some assumed max sequence length $L \in \mathbb{N}^+$, where only the first $l$ columns of $W^{position}$ are used. $LN$ refers to the LayerNorm function \citep{ba2016layernormalization}.

The major component of transformers are the \textit{self-attention} mechanism, which generates weighted combinations of the sequence $\bar{\ovec}$, $f(\bar{\ovec}) = \text{softmax}( \epsilon {QK^T}) V.$
We call $Q = \bar{\ovec}W^{Q}$, $V = \bar{\ovec}W^{V}$, and $K = \bar{\ovec}W^{K}$ the query, key, and value, respectively, and $\epsilon = 1 / \sqrt{h}$ is a constant chosen to prevent the dot products from causing extremely peaked softmax distributions. The softmax operator, which converts vectors of real numbers to vectors of probabilities, $\text{softmax}(\ovec)_i = \exp{(\ovec_i)} /  \sum^l_{j=1} \exp(o_j)$, 
defines the weight each vector $\ovec_i$ contributes. The parameter set $W^{attn} = \{W^{Q}, W^{V}, W^{K}\} \in \{\mathbb{R}^{h \times h}, \mathbb{R}^{h \times h}, \mathbb{R}^{h \times h} \}$ are linear projections. We learn model parameters with gradient descent. Self-attention is followed by a nonlinear transformation function $f(\bar{\ovec})$ and residual connection to form transformer layer $T_i(\bar{\ovec}) =  W^{out} \sigma(W^{in}(\text{LN}(\bar{\ovec} + f(\bar{\ovec}))) + \text{LN}(f(\bar{\ovec})) + \bar{\ovec}$, where $W^{out}, W^{in} \in \mathbb{R}^{h \times h}$, and $\sigma$ are ReLU functions.

\section{Efficient Morphology Transfer Learning}
\label{sec:methodology}

This section discusses our work investigating the efficacy of PEFT algorithms for morphology-aware online RL. Control policies can require immense computation and physical resources to learn for real world systems (e.g. robotics). If policies can explicitly account for agent morphology, this can reduce computation costs by aggregating knowledge between morphologies and improve the policy's generalization capabilities to new embodiments. 

Unfortunately, a morphology-aware policy may not elicit the optimal performance of a target morphology due to these generalization capabilities. For real-world applications, it is likely necessary that pretrained model components continue to learn to maximize task performance. Reducing the total necessary learnable parameters is thus significant to achieving this result because, at deployment, it may not be feasible to access sufficient computation resources to perform learning updates. These limitations motivate the potential of PEFT solutions, which are applicable in varying resource limitations when deploying these policies.

We formalize the parameter efficient finetuning problem by first assuming access to a trained policy $\pi(\svec, \cvec ~; \theta^\star)$ with optimized parameters $\theta^\star$. For a new morphology $\bar{c} \sim p(C)$, we expect this policy to have some initial performance $\mathbb{E}_{\pi(s, \bar{c}; \theta^\star)}[G_{\bar{c}}]$ which we call the \textit{zero-shot} policy performance. The goal of our work is to optimize new parameters $\phi$ to maximize the cumulative reward objective, 
\begin{equation*}
   \phi^{\star} = \argmax_\phi \mathbb{E}_{\pi_{}(s, \bar{c}~; \theta^{\star} \cup \phi)}[G_{\bar{c}}(s)],
\end{equation*}
where the new parameters are optimized only for the specific morphology $\bar{c}$. We hypothesize that learning a small set $\phi$ will perform measurably better than the base policy's zero-shot performance,  $\mathbb{E}_{\pi(s, \bar{c}~;\theta^{\star} \cup \phi^{\star})}[G_{\bar{c}}(s)] > \mathbb{E}_{\pi(s, \bar{c}~; \theta^{\star})}[G_{\bar{c}}(s)]$ where  $|\phi| \ll |\theta|$. We investigate this problem learning policies using the \textit{Metamorph} framework and then finetuning this pre-trained policy with one of several PEFT techniques, which describe next. 

\subsection{Metamorph Framework}

Metamorph is morphology-aware learning framework that is an instantiation of the CMDP formulation we described in Section~\ref{sec:background}. In Metamorph, a policy is trained over a set of 100 training morphologies.\footnote{We explicitly mention training on 100 morphologies because that is done in the original paper. Any number of training morphologies can be used in practice.}
Each morphology $c$ induces an observation sequence  $\ovec = [\ovec_1, \ovec_2, \ovec_3, ..., \ovec_{l(c)}]$ \textit{for each} time step. 
To account for varying $l(c) \in \mathbb{N}^+$ between morphologies the policy is a transformer (Section~\ref{sec:background}). The transformer encoders hidden representations $\hvec \in \mathbb{R}^{l(c) \times h}$ with $h \in \mathbb{N}^+$ hidden features per limb. Actions are predicted with a multi-layer perceptron \textit{per limb} as $\avec_i = g_\theta(\hvec_i)$, where $g: \mathcal{H} \rightarrow \mathcal{A}$ is a mapping from hidden representations to actions. Here, $\avec_i \in \mathbb{R}$ while $\hvec_i \in \mathbb{R}^h$. Having a token per limb enables a metamorph policy to adapt to varying limb configurations in practice. The policy $\pi_\theta(\ovec)$ is optimized using Proximal Policy Optimization ~\citep{schulman2017PPO}. 

We chose to use this framework because it uses transformer-based policies as the morphology-aware policy. Several PEFT techniques we consider in this paper are designed specifically for use with transformer models. The framework code is open sourced, making it accessible to researchers to reproduce our results and compare other PEFT techniques in potential future work. Several works have also built off this repository to improve the base-architecture design \citep{xiong2023modumorph,xiong2024distillingHypernetworks}.  

\begin{figure}[h]
    \centering
    \includegraphics[trim={0mm 0 0 0},clip,width=0.7\textwidth]{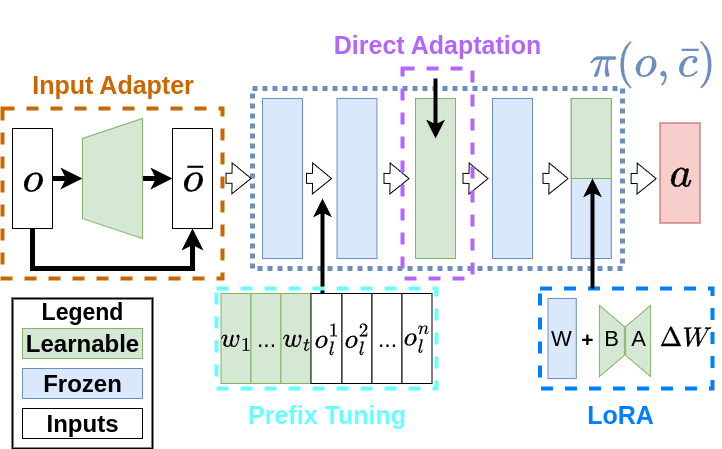}
    \caption{ A visualization of the various PEFT techniques considered in this paper. We investigate applying PEFT techniques \textit{independently} from each other.  
    }
    \label{fig:peft-techniques}
\end{figure}

\subsection{Parameter Efficient Finetuning Across Morphologies}

We group PEFT approaches as either direct, input, or prefix adaptation techniques. \textit{Direct adaptive} PEFT approaches modify some subset of the weights $\phi \subseteq \theta^{\star}$ or else add learnable delta weights $\hat{W} = W + \Delta W$. \textit{Input-adaptive} PEFT approaches perform some transformation of the inputs to elicit the optimal performance in the model. Prefix tuning prepends a learnable sequenes of tokens to each input sequence. We visualize the various types of PEFT algorithms evaluated in Figure~\ref{fig:peft-techniques}.

We consider tuning subsets of $\theta^{\star}$ for direct adaptive PEFT learning. We list the combinations studied and their experiment result identifiers in Appendix~\ref{sec:appendix_directfinetuning}, for example \textit{Layer 5} represents directly tuning the final transformer layer. For attention and nonlinear transformer layers, we used low-lank adapters (LoRA) \citep{HuSWALWWC22Lora}, to learn $\Delta W \in \mathbb{R}^{h^1 \times h^2} = AB$, where $A \in \mathbb{R}^{h^1 \times r}$ and $B \in \mathbb{R}^{r \times h^2}$ are low-rank matrices of rank $r \in \mathbb{N}^+$ to reduce learnable weights for the weight dimensions $h_1 \in \mathbb{N}^+, h_2 \in \mathbb{N}^+$. We describe LoRA initialization details in the Appendix~\ref{sec:appendix_lora}.  


For input-adaptive PEFT approaches, we consider learning an extra input adapter layer. We consider an input adapter layer that modifies the policy observation as $h: \mathbb{R}^{d^c} \rightarrow \mathbb{R}^{d^c}$, so that policy uses modified inputs $a \sim \pi_{\theta^{\star}}(h(\ovec))$. We consider two variations of the function $h$ where one is a direct nonlinear transform $h(o) = H^{out}\sigma(H^{in}\ovec)$ or else a nonlinear transformation with a residual connection $h(\ovec) = \ovec + H^{out}\sigma(H^{in}\ovec)$, with learnable weights $\phi =\{H^{in}, H^{out} \}$. We use a hidden layer size of 256 units. The input adapter transforms observations to elicit better performance from a frozen pre-trained model.


\textit{Prefix-tuning} is a PEFT approach where a set of learnable tokens are pre-pended to the input sequence to elicit desired outputs from the model~\citep{LiL20Prefix-tuning}. These prefixes are a sequence $\phi = [\wvec_{1}, \wvec_{ 2}, ..., \wvec_{m} ]$ of $m \in \mathbb{N}^+$ tokens, where $\wvec_i \in \mathbb{R}^{h}$ is a vector. These tokens are then pre-pended to the observations $o^{\textit{prefix}} = [\phi; \ovec_1, \ovec_2, ..., \ovec_{l(c)}]$ and otherwise processed normally by the transformer layers. Tokens optionally can be pre-pended deeper in the model (e.g., $\ovec^{\textit{prefix}}_l = [\phi; T^l(\ovec^{l-1})]$ for layer $l > 1$) or multiples prefix sets can be used (e.g., $\phi = \{\phi_1, \phi_2, ..., \phi_l\}$ would be learnable prefixes for each layer). 

\section{Experiments}
\label{sec:experiments}

\begin{figure}[h]
    \centering
    \begin{subfigure}[b]{0.3\textwidth}
        \centering
        \includegraphics[width=\textwidth]{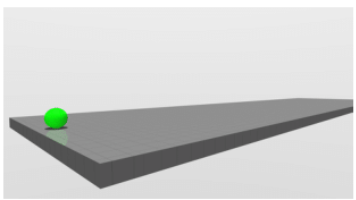}
        \caption{Flat Terrain }
    \end{subfigure}
    \begin{subfigure}[b]{0.3\textwidth}
        \centering
        \includegraphics[width=\textwidth]{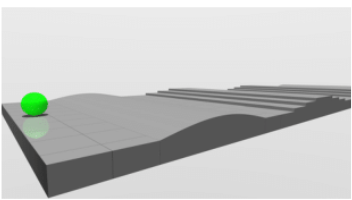}
        \caption{Variable Terrain}
    \end{subfigure}
    \begin{subfigure}[b]{0.3\textwidth}
        \centering
        \includegraphics[width=\textwidth]{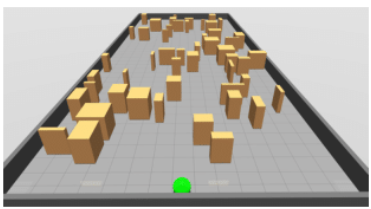}
        \caption{Obstacle Avoidance}
    \end{subfigure}
    \caption{Locomotion environments. Diagrams are reproduced from \cite{gupta2022metamorph}.}
    \label{fig:environments}
\end{figure}
\begin{figure*}[h]
    \vspace{5mm}
    \centering
    
    \begin{subfigure}[b]{0.16\textwidth}
        \includegraphics[width=\textwidth]{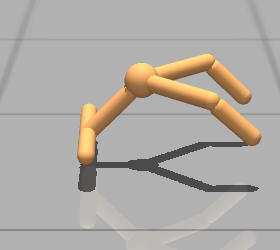}
        \caption*{Morphology 1}
    \end{subfigure}
    \hfill
    \begin{subfigure}[b]{0.16\textwidth}
        \includegraphics[width=\textwidth]{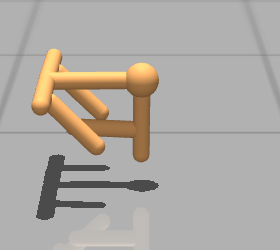}
        \caption*{Morphology 2$^{*}$}
    \end{subfigure}
    \hfill
    \begin{subfigure}[b]{0.16\textwidth}
        \includegraphics[width=\textwidth]{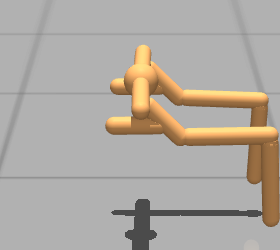}
        \caption*{Morphology 3}
    \end{subfigure}
    \hfill
    \begin{subfigure}[b]{0.16\textwidth}
        \includegraphics[width=\textwidth]{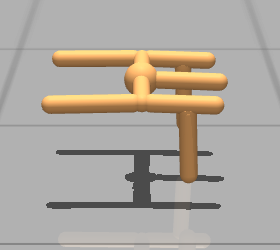}
        \caption*{Morphology 4$^{**}$}
    \end{subfigure}
    \hfill
    \begin{subfigure}[b]{0.16\textwidth}
        \includegraphics[width=\textwidth]{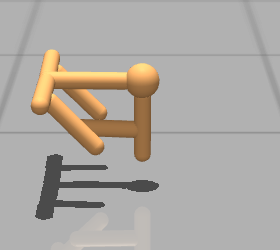}
        \caption*{Morphology 5$^{*}$}
    \end{subfigure}
    \hfill
    \begin{subfigure}[b]{0.16\textwidth}
        \includegraphics[width=\textwidth]{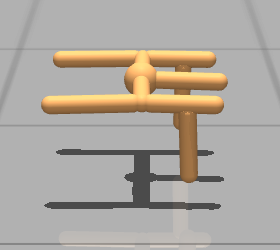}
        \caption*{Morphology 6$^{**}$}
        
    \end{subfigure}
    \caption{The six testing morphologies used in our evaluation. Morphology numbers correspond to those shown in relevant results. Morphologies \{2, 4\}$^{*}$  and \{4, 6\}$^{**}$ have the same limb configurations but different kinematic and dynamic parameters.  } 
    \label{fig:morphologies}
\end{figure*}


This research aims to evaluate the efficacy of PEFT approaches for online learning on target morphologies. These experiments strive to address the following research questions: \textit{(1) How effectively does each PEFT learning approach compare between each other and end-to-end finetuning? \newline(2) What is the relationship between the total number of learnable parameters and the performance when adapting to target morphologies? (3) What are the relevant factors for using prefix tuning and LoRA in online reinforcement learning?} Our results contribute to understanding the efficacy of these approaches in online learning, and can help guide future research developing PEFT algorithms for morphology-aware policy transfer. As part of our experiments, we also compare to learning a policy from scratch to determine whether or not if pretraining does help policy transfer. 

We report experimental findings on the efficacy of different forms of parameter-efficient finetuning in morphological transfer. We use three locomotion tasks that differ in the terrain types shown in Figure~\ref{fig:environments}; these include a flat surface, randomized variable terrain, and rectangular obstacle avoidance. Each task's reward function emphasizes running as fast as possible to the right. To evaluate the PEFT techniques, we randomly sampled six morphologies from the Metamorph test dataset \citep{gupta2022metamorph}. We visualize the testing morphologies in Figure~\ref{fig:morphologies} which include four unique limb configurations and two sets of varying kinematic and dynamic differences. We evaluate PEFT techniques on eighteen environment-morphology combinations. 

As mentioned in Section~\ref{sec:methodology}, we generate our pre-trained models using the Metamorph framework with default hyperparameters \citep{gupta2022metamorph}. We train five base models using one hundred training morphologies for ten million time steps for each environment. 
We then apply each PEFT approach with the pre-trained models on the six test morphologies for five million timesteps each. We repeat experiments for five random seeds for every set of PEFT hyperparameters we report. For each seed, we use one of the pre-trained models without replacement. We use the same learning hyperparameters for the pre-training phase, except we \textit{do not use Dropout} in the transformer embedding. Previous research shows that Dropout is helpful for Metamorph pre-training \citep{xiong2023modumorph}, but in preliminary evaluations, we found Dropout was not helpful for finetuning models.

\subsection{Best Performances Across Methods}

\begin{figure}[h]
    \centering
    \vspace{5mm}
    
    \includegraphics[width=\linewidth]{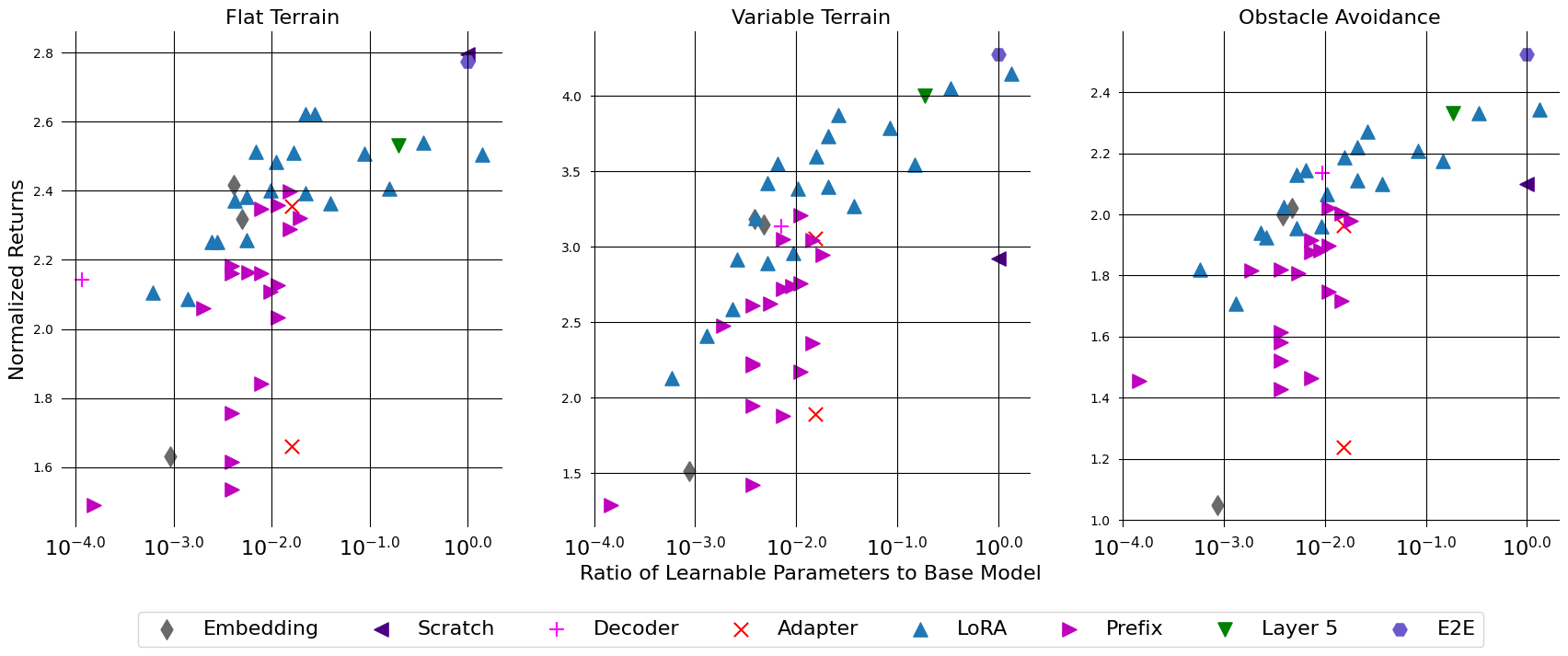}
    
    \caption{Percentage of trainable ratios to total base model parameters vs achieved normalized results. Results suggest total learnable parameters are a contributing factor in final policy performance.}
    \label{fig:param-to-perf}
\end{figure}

In this section, we report results towards answering our first two research questions on the efficacy of different PEFT techniques. We report results in Figure~\ref{fig:param-to-perf} which shows the performance of different PEFT techniques. We normalize cumulative rewards after performing parameter-efficient finetuning by the pre-trained policy's zero-shot performance averaging these scores across the six testing morphologies. The x-axis shows percentage of learnable parameters to the base-models original parameter counts. We include the original cumulative reward scores by best PEFT hyperparameter configuration in Appendix~\ref{sec:appendix_best PEFT}. 

Our results reveal a number of notable trends across PEFT approaches. An interesting finding suggests that morphology-pretraining utility is dependent on task complexity. On the flat terrain tasks, learning from scratch is comparable to end-to-end finetuning but between variable terrain or obstacle avoidance learning-from-scratch performs substantially worse. Across morphologies, results suggest that the best input-learnable configurations behave similarly to directly tuning the input Embedding and Decoder, suggesting some equivalence between the two approaches for the model sizes used in our experiments. We observed substantial performance improvements tuning just the fifth transformer block, suggesting that if direct model access is possible and a more generous computation budget is available, this layer substantially influences the policy performance. When possible, results suggest more learning parameters are generally favorable given end-to-end finetuning results.

\subsection{Ablation of LoRA and Prefix Tuning}

\begin{figure}[h!]
    \centering
    \begin{subfigure}[b]{0.48\textwidth}
        \centering
        \includegraphics[width=.8\textwidth]{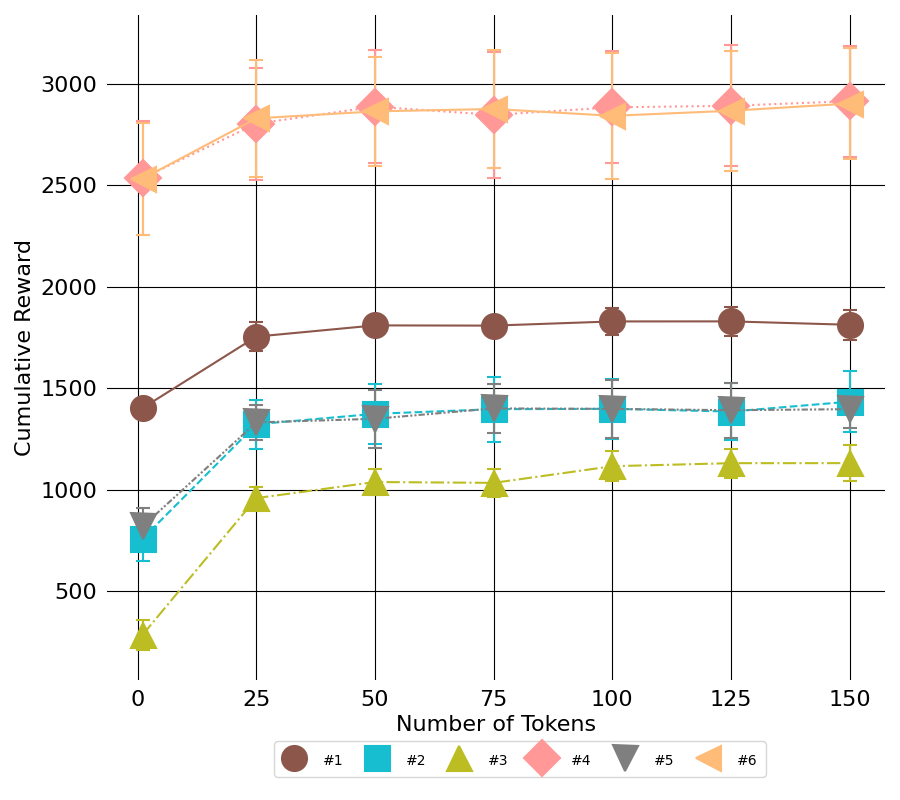}
        \caption{Number of randomly initialized prefix tokens} 
        \label{fig:ablation-n-tokens}
    \end{subfigure}
    \hfill
    \begin{subfigure}[b]{0.48\textwidth}
        \centering
        \includegraphics[width=.85\textwidth]{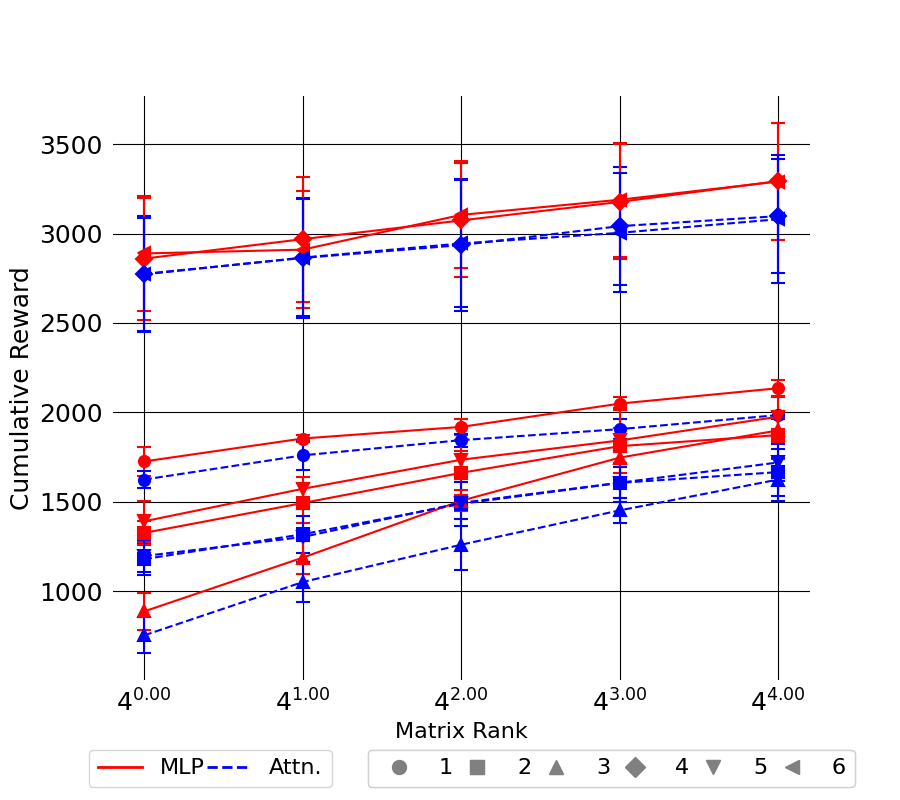}
        \caption{Lora in different layers of fifth transformer block. }
        \label{fig:ablation-lora-layer}
    \end{subfigure}
    \caption{Ablation studies on prefix tokens and LoRA  in variable terrain. } 
    \label{fig:combined-ablation}
\end{figure}

In this section, we report results comparing different hyperparameter choices for LoRA and Prefix approaches to address our third research question. We include additional results in Appendix~\ref{sec:appendix_prefixadditional}. 
The reported results represent the consistent behaviours observed between the evaluations in each environment. Figure~\ref{fig:ablation-lora-layer} shows the results of using LoRA in either the nonlinear transformations (MLP) or attention layer (Attn.) of the fifth transformer layer. The results show that across morphologies for a single layer's full rank matrices are necessary. Applying LoRA to the nonlinear transformation is preferable for adaption to elicit optimal performance, but results suggest that directly tuning a single layer can be better to avoid introducing more learning parameters.

Prefixing tuning results have more nuanced conclusions. We consider three major factors for effective prefix usage: (1) the number of tokens, (2) the injection layer, and (3) comparing token initialization approaches. Each factor has been shown to substantially impact performance \citep{Ding2023PEFTLscaleLM,LiL20Prefix-tuning}. For (3), we propose a second pretraining stage to learn morphology-aware tokens. This second stage repeats the Metamorph training but keeps the base model frozen while learning the tokens.   

\begin{figure}[h!]
    \vspace{5mm}
    \centering
    \includegraphics[width=.8\textwidth]{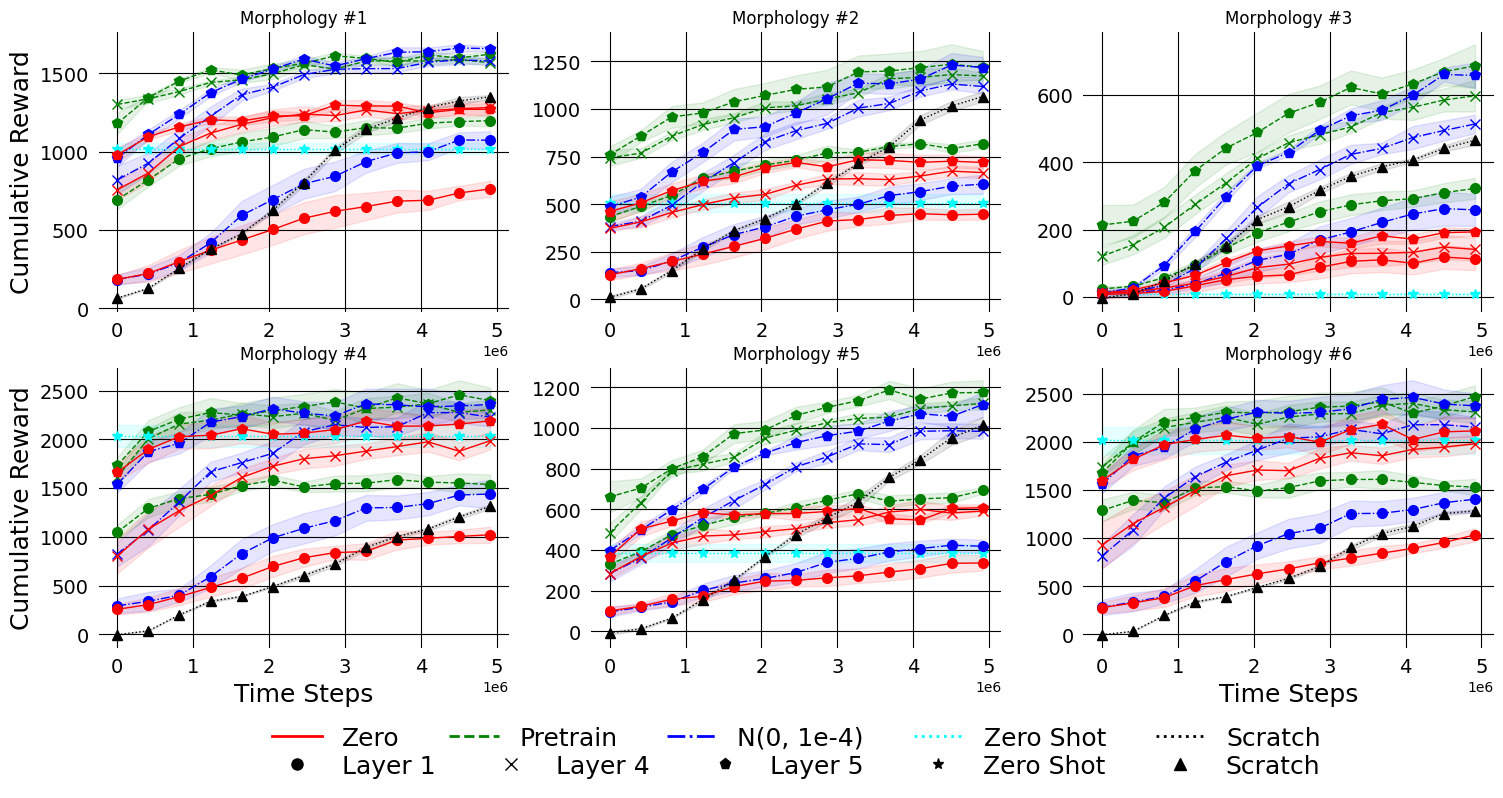}
    \caption{Choice of initialization and injection layers of prefix tuning in variable terrain.  Initial zero-shot results of E2E learning are plotted to compare affect of prefixes.}
    \label{fig:prefix-injection-layer-ablation}
\end{figure}

We generally observe that more learnable parameters are beneficial, such as by increasing the number of tokens used (see Figure~\ref{fig:ablation-n-tokens}), which agrees with our other findings previously discussed. In our experiments, a complication with prefix tuning is that introducing un-trained tokens can negatively impact policy zero-shot performance. This performance drop can occur because the base model is not trained jointly with the prefix which initially are noisy observations. This problem is largely missed in supervised learning applications because performance is evaluated \textit{after training}.
In contrast, we care for performance \textit{during training} especially because it's preferable policies have strong initial performance for real-world systems to avoid consequences of poor-performing policies (e.g., damage to the hardware). We conducted experiments adding $50$ prefix tokens as input before different transformer blocks to investigate their impact on learning performance. We compared different token initializations, including zero vectors, small Gaussian noise ($N(0; 10^{-4})$), or the pretrained tokens as previously described in this section. We include results when learning from scratch to understand the value of pretraining for sample efficiency. Figure~\ref{fig:prefix-injection-layer-ablation} show's learning curves.  

Generally, we observed that the initial zero-shot performance is often negatively affected by zero or random initialization approaches, especially when introducing prefix tokens to the earlier transformer layers. This result suggests that deep layers are less sensitive to the base models' perturbations and better steer feature representations for target morphologies. Interestingly, pre-trained prompting embeddings significantly improved policy performance during learning compared to other initialization approaches, especially on Morphology \#3, which we found most PEFT approaches struggled to learn. This demonstrates that prefix initialization can mitigate loss in zero-shot performance during finetuning in online learning.

\section{Conclusions and Future Work}

In this paper, we have investigated the impact of PEFT approaches for finetuning morphology-aware policies. We demonstrate that in most cases, one should train as many parameters online as possible to elicit the best performances of a pre-trained policy. Our analysis reveals that many PEFT approaches provide substantial benefits in deeper layers, so tuning the final transformer block is likely effective for policy finetuning. In scenarios where directly finetuning the base model is difficult, learnable inputs perform similarly to tuning either the input embeddings or decoder layers of the transformer-based policy.

There are several promising future research directions to extend our findings. One crucial factor, particularly for prefix tuning, is the scale of the model. Many reported successes of PEFT approaches are on models with tens of millions to billions of parameters \citep{LiL20Prefix-tuning}. In this work, we used relatively small models ($\sim$3.5 million parameters at most between policy and value function in PPO). We also focused on vanilla transformer architectures used in Metamorph, but researchers have proposed variations for morphology-aware policies \citep{trabucco2022anymorph,xiong2023modumorph}. Given the promise of PEFT techniques in RL, we see much potential for future development in PEFT development for online learning.


\subsubsection*{Broader Impact Statement}
\label{sec:broaderImpact}
Although our work focuses on the positives of input adapter and prefix tuning techniques, an undesirable consequences of this work is revealing a means of adversarially attacking deep learning-based control policies. In AI security research this is called \textit{adversarial 
 reprogramming} in which models are repurposed for nefarious uses  \citep{elsayed2018adversarial,yang2023WhyAdvReprograWorksWhenfail,englert2022adversarial}. Our results show that input adapter finetuning approaches can measurably affect policy performance without changing the base policy weights. We hypothesize that these techniques could be used to re-purpose pre-trained policies for other tasks, potentially in adversarial ways. Some scenarios we imagine this could arise is through benign adversarial action decisions by the pre-trained policy (delaying purchase in investment agent systems, adding extra torque during control, etc.). Given these implications, we caution that research in PEFT techniques should also consider the negative consequence of using input adapter or prefix tuning approaches for control tasks.  

 \subsubsection*{Acknowledgments}
\label{sec:ack}

This work has taken place in part in the Intelligent Robot Learning (IRL) Lab and Vision \& Robotics Lab at the University of Alberta, which are supported in part by research grants from Alberta Innovates; Alberta Machine Intelligence Institute (Amii); a Canada CIFAR AI Chair, Amii; Digital Research Alliance of Canada; Mitacs; and the National Science and Engineering Research Council (NSERC). We further acknowledge working with members of the Robotics and Embodied AI Lab (REAL) at the Université de Montréal and Mila who received funding support from the Natural Sciences and Engineering Research Council (NSERC) of Canada, Samsung AI Lab, Google Research,  the Canadian Institute for Advanced Research (CIFAR) and compute support from Digital Research Alliance of Canada, Mila IDT, and NVidia. We thank Intel Labs for funding this research. 
\appendix


\bibliography{bibliography_cleaned}

\begin{thebibliography}{51}
\providecommand{\natexlab}[1]{#1}
\providecommand{\url}[1]{\texttt{#1}}
\expandafter\ifx\csname urlstyle\endcsname\relax
  \providecommand{\doi}[1]{DOI: #1}\else
  \providecommand{\doi}{DOI: \begingroup \urlstyle{rm}\Url}\fi

\bibitem[Ba et~al.(2016)Ba, Kiros, and Hinton]{ba2016layernormalization}
Jimmy~Lei Ba, Jamie~Ryan Kiros, and Geoffrey~E. Hinton.
\newblock Layer normalization.
\newblock \emph{CoRR}, abs/1607.06450, 2016.

\bibitem[Bohlinger et~al.(2025)Bohlinger, Czechmanowski, Krupka, Kicki, Walas, Peters, and Tateo]{bohlinger2025onepolicytorun}
Nico Bohlinger, Grzegorz Czechmanowski, Maciej~Piotr Krupka, Piotr Kicki, Krzysztof Walas, Jan Peters, and Davide Tateo.
\newblock One policy to run them all: an end-to-end learning approach to multi-embodiment locomotion.
\newblock In \emph{Conference on Robot Learning}, 2025.

\bibitem[Bousmalis et~al.(2024)]{bousmalis2024robocat}
Konstantinos Bousmalis et~al.
\newblock {RoboCat}: A self-improving generalist agent for robotic manipulation.
\newblock \emph{Transactions on Machine Learning Research}, 2024.

\bibitem[Brunskill \& Li(2013)Brunskill and Li]{brunskill2013sample}
Emma Brunskill and Lihong Li.
\newblock Sample complexity of multi-task reinforcement learning.
\newblock In \emph{Conference on Uncertainty in Artificial Intelligence}, 2013.

\bibitem[Chen et~al.(2023)Chen, Han, Sun, and Huang]{chen2023subequivariant}
Runfa Chen, Jiaqi Han, Fuchun Sun, and Wenbing Huang.
\newblock Subequivariant graph reinforcement learning in 3{D} environments.
\newblock In \emph{International Conference on Machine Learning}, 2023.

\bibitem[Chen et~al.(2018)Chen, Murali, and Gupta]{ChenMG18Hardware}
Tao Chen, Adithyavairavan Murali, and Abhinav Gupta.
\newblock Hardware conditioned policies for multi-robot transfer learning.
\newblock In \emph{Advances in Neural Information Processing Systems}, 2018.

\bibitem[Deng et~al.(2023)Deng, Gu, Zheng, Chen, Stevens, Wang, Sun, and Su]{deng2023mind2web}
Xiang Deng, Yu~Gu, Boyuan Zheng, Shijie Chen, Samuel Stevens, Boshi Wang, Huan Sun, and Yu~Su.
\newblock {MIND2WEB}: towards a generalist agent for the web.
\newblock In \emph{Advances in Neural Information Processing Systems}, 2023.

\bibitem[D'Eramo et~al.(2020)D'Eramo, Tateo, Bonarini, Restelli, and Peters]{deramo2020sharing}
C~D'Eramo, D~Tateo, A~Bonarini, M~Restelli, and J~Peters.
\newblock Sharing knowledge in multi-task deep reinforcement learning.
\newblock In \emph{International Conference on Learning Representations}, 2020.

\bibitem[Ding et~al.(2023)Ding, Qin, Yang, Wei, Yang, Su, Hu, Chen, Chan, Chen, Yi, Zhao, Wang, Liu, Zheng, Chen, Liu, Tang, Li, and Sun]{Ding2023PEFTLscaleLM}
Ning Ding, Yujia Qin, Guang Yang, Fuchao Wei, Zonghan Yang, Yusheng Su, Shengding Hu, Yulin Chen, Chi-Min Chan, Weize Chen, Jing Yi, Weilin Zhao, Xiaozhi Wang, Zhiyuan Liu, Hai-Tao Zheng, Jianfei Chen, Yang Liu, Jie Tang, Juanzi Li, and Maosong Sun.
\newblock Parameter-efficient fine-tuning of large-scale pre-trained language models.
\newblock \emph{Nature Machine Intelligence}, 5\penalty0 (3):\penalty0 220--235, 2023.

\bibitem[Dong et~al.(2023)Dong, Li, Dai, Zheng, Wu, Chang, Sun, Xu, Li, and Sui]{Dong23IncontextSurvey}
Qingxiu Dong, Lei Li, Damai Dai, Ce~Zheng, Zhiyong Wu, Baobao Chang, Xu~Sun, Jingjing Xu, Lei Li, and Zhifang Sui.
\newblock A survey for in-context learning.
\newblock \emph{CoRR}, abs/2301.00234, 2023.

\bibitem[Driess et~al.(2023)Driess, Xia, Sajjadi, Lynch, Chowdhery, Ichter, Wahid, Tompson, Vuong, Yu, et~al.]{driess2023palm}
Danny Driess, Fei Xia, Mehdi~SM Sajjadi, Corey Lynch, Aakanksha Chowdhery, Brian Ichter, Ayzaan Wahid, Jonathan Tompson, Quan Vuong, Tianhe Yu, et~al.
\newblock {PaLM-E}: An embodied multimodal language model.
\newblock In \emph{International Conference on Machine Learning}, 2023.

\bibitem[Du et~al.(2020)Du, Kakade, Wang, and Yang]{Du2020IsGoodRepresentationSufficient}
Simon~S. Du, Sham~M. Kakade, Ruosong Wang, and Lin~F. Yang.
\newblock Is a good representation sufficient for sample efficient reinforcement learning?
\newblock In \emph{International Conference on Learning Representations}, 2020.

\bibitem[Elsayed et~al.(2019)Elsayed, Goodfellow, and Sohl-Dickstein]{elsayed2018adversarial}
Gamaleldin~F. Elsayed, Ian Goodfellow, and Jascha Sohl-Dickstein.
\newblock Adversarial reprogramming of neural networks.
\newblock In \emph{International Conference on Learning Representations}, 2019.

\bibitem[Englert \& Lazic(2022)Englert and Lazic]{englert2022adversarial}
Matthias Englert and Ranko Lazic.
\newblock Adversarial reprogramming revisited.
\newblock In \emph{Advances in Neural Information Processing Systems}, 2022.

\bibitem[Furuta et~al.(2023)Furuta, Iwasawa, Matsuo, and Gu]{furuta2023mxtbench}
Hiroki Furuta, Yusuke Iwasawa, Yutaka Matsuo, and Shixiang~Shane Gu.
\newblock A system for morphology-task generalization via unified representation and behavior distillation.
\newblock In \emph{International Conference on Learning Representations}, 2023.

\bibitem[Gupta et~al.(2022)Gupta, Fan, Ganguli, and Fei-Fei]{gupta2022metamorph}
Agrim Gupta, Linxi Fan, Surya Ganguli, and Li~Fei-Fei.
\newblock {MetaMorph}: Learning universal controllers with transformers.
\newblock In \emph{International Conference on Learning Representations}, 2022.

\bibitem[Hallak et~al.(2015)Hallak, Castro, and Mannor]{hallak2015contextualmarkovdecisionprocesses}
Assaf Hallak, Dotan~Di Castro, and Shie Mannor.
\newblock Contextual {Markov} decision processes.
\newblock \emph{CoRR}, abs/1502.02259, 2015.

\bibitem[Henderson et~al.(2018)Henderson, Islam, Bachman, Pineau, Precup, and Meger]{henderson2018deepRLthatMatters}
Peter Henderson, Riashat Islam, Philip Bachman, Joelle Pineau, Doina Precup, and David Meger.
\newblock Deep reinforcement learning that matters.
\newblock In \emph{AAAI Conference on Artificial Intelligence}, 2018.

\bibitem[Hong et~al.(2022)Hong, Yoon, and Kim]{Hong2022SWAT}
Sunghoon Hong, Deunsol Yoon, and Kee{-}Eung Kim.
\newblock Structure-aware transformer policy for inhomogeneous multi-task reinforcement learning.
\newblock In \emph{International Conference on Learning Representations}, 2022.

\bibitem[Hu et~al.(2022)Hu, Shen, Wallis, Allen{-}Zhu, Li, Wang, Wang, and Chen]{HuSWALWWC22Lora}
Edward~J. Hu, Yelong Shen, Phillip Wallis, Zeyuan Allen{-}Zhu, Yuanzhi Li, Shean Wang, Lu~Wang, and Weizhu Chen.
\newblock {LoRA}: Low-rank adaptation of large language models.
\newblock In \emph{International Conference on Learning Representations}, 2022.

\bibitem[Huai et~al.(2019)Huai, Ding, Wang, Geng, and Zhang]{huai2019towardsDLresourceconstraint}
Zhibo Huai, Bo~Ding, Huaimin Wang, Mingyang Geng, and Lei Zhang.
\newblock Towards deep learning on resource-constrained robots: A crowdsourcing approach with model partition.
\newblock In \emph{IEEE SmartWorld, Ubiquitous Intelligence Computing, Advanced Trusted Computing, Scalable Computing Communications, Cloud Big Data Computing, Internet of People and Smart City Innovation}, 2019.

\bibitem[Huang et~al.(2020)Huang, Mordatch, and Pathak]{HuangMP20Onepolicy}
Wenlong Huang, Igor Mordatch, and Deepak Pathak.
\newblock One policy to control them all: Shared modular policies for agent-agnostic control.
\newblock In \emph{International Conference on Machine Learning}, 2020.

\bibitem[Kirk et~al.(2023)Kirk, Zhang, Grefenstette, and Rockt\"{a}schel]{kirk2023SurveyzeroshotRL}
Robert Kirk, Amy Zhang, Edward Grefenstette, and Tim Rockt\"{a}schel.
\newblock A survey of zero-shot generalisation in deep reinforcement learning.
\newblock \emph{Journal of Artificial Intelligence Research}, 76, 2023.

\bibitem[Kurin et~al.(2021)Kurin, Igl, Rockt{\"{a}}schel, Boehmer, and Whiteson]{KurinIRBW21Mybody}
Vitaly Kurin, Maximilian Igl, Tim Rockt{\"{a}}schel, Wendelin Boehmer, and Shimon Whiteson.
\newblock My body is a cage: the role of morphology in graph-based incompatible control.
\newblock In \emph{International Conference on Learning Representations}, 2021.

\bibitem[Lee et~al.(2023)Lee, Chen, Tajwar, Kumar, Yao, Liang, and Finn]{lee2022surgical}
Yoonho Lee, Annie~S Chen, Fahim Tajwar, Ananya Kumar, Huaxiu Yao, Percy Liang, and Chelsea Finn.
\newblock Surgical fine-tuning improves adaptation to distribution shifts.
\newblock In \emph{International Conference on Learning Representations}, 2023.

\bibitem[Lester et~al.(2021)Lester, Al-Rfou, and Constant]{lester2021powerscaleforPEFT}
Brian Lester, Rami Al-Rfou, and Noah Constant.
\newblock The power of scale for parameter-efficient prompt tuning.
\newblock In \emph{Conference on Empirical Methods in Natural Language Processing}, 2021.

\bibitem[Li et~al.(2024)Li, Li, Zhu, and Zhao]{LiLZZ24MAT}
Boyu Li, Haoran Li, Yuanheng Zhu, and Dongbin Zhao.
\newblock {MAT}: Morphological adaptive transformer for universal morphology policy learning.
\newblock \emph{IEEE Transactions on Cognitive and Developmental Systems}, 16\penalty0 (4):\penalty0 1611--1621, 2024.

\bibitem[Li \& Liang(2021)Li and Liang]{LiL20Prefix-tuning}
Xiang~Lisa Li and Percy Liang.
\newblock Prefix-tuning: Optimizing continuous prompts for generation.
\newblock In \emph{Annual Meeting of the Association for Computational Linguistics}, 2021.

\bibitem[Liu et~al.(2022)Liu, Ji, Fu, Tam, Du, Yang, and Tang]{liu2022ptuningprmopting}
Xiao Liu, Kaixuan Ji, Yicheng Fu, Weng Tam, Zhengxiao Du, Zhilin Yang, and Jie Tang.
\newblock {P}-tuning: Prompt tuning can be comparable to fine-tuning across scales and tasks.
\newblock In \emph{Annual Meeting of the Association for Computational Linguistics}, 2022.

\bibitem[Liu et~al.(2024)Liu, Zhang, Asadi, Liu, Zhao, Sabach, and Fakoor]{LiuZA0ZSF24TAIL}
Zuxin Liu, Jesse Zhang, Kavosh Asadi, Yao Liu, Ding Zhao, Shoham Sabach, and Rasool Fakoor.
\newblock {TAIL}: Task-specific adapters for imitation learning with large pretrained models.
\newblock In \emph{International Conference on Learning Representations}, 2024.

\bibitem[Luck et~al.(2020)Luck, Amor, and Calandra]{luck2020DataeffCoadptmorphandBehavior}
Kevin~Sebastian Luck, Heni~Ben Amor, and Roberto Calandra.
\newblock Data-efficient co-adaptation of morphology and behaviour with deep reinforcement learning.
\newblock In \emph{Conference on Robot Learning}, 2020.

\bibitem[Maurer et~al.(2016)Maurer, Pontil, and Romera-Paredes]{maurer2016benefit}
Andreas Maurer, Massimiliano Pontil, and Bernardino Romera-Paredes.
\newblock The benefit of multitask representation learning.
\newblock \emph{Journal of Machine Learning Research}, 17\penalty0 (81):\penalty0 1--32, 2016.

\bibitem[Neuman et~al.(2022)Neuman, Plancher, Duisterhof, Krishnan, Banbury, Mazumder, Prakash, Jabbour, Faust, de~Croon, and Reddi]{neuman2022TinyRobotLearning}
Sabrina~M. Neuman, Brian Plancher, Bardienus~P. Duisterhof, Srivatsan Krishnan, Colby Banbury, Mark Mazumder, Shvetank Prakash, Jason Jabbour, Aleksandra Faust, Guido~C.H.E. de~Croon, and Vijay~Janapa Reddi.
\newblock Tiny robot learning: Challenges and directions for machine learning in resource-constrained robots.
\newblock In \emph{IEEE International Conference on Artificial Intelligence Circuits and Systems}, 2022.

\bibitem[Nie et~al.(2023)Nie, Ni, Chang, Meng, Huo, Xiang, and Tian]{nie2023protuningUnifiedvision}
Xing Nie, Bolin Ni, Jianlong Chang, Gaofeng Meng, Chunlei Huo, Shiming Xiang, and Qi~Tian.
\newblock Pro-tuning: Unified prompt tuning for vision tasks.
\newblock \emph{IEEE Transactions on Circuits and Systems for Video Technology}, 34\penalty0 (6):\penalty0 4653--4667, 2023.

\bibitem[{Octo Model Team}(2024)]{octoteam2024OctoOpensourcegeneralist}
{Octo Model Team}.
\newblock Octo: An open-source generalist robot policy.
\newblock In \emph{Robotics: Science and Systems}, 2024.

\bibitem[{Open X-Embodiment Team}(2024)]{ONeillRMGPLPGMJ24Embodiment-X}
{Open X-Embodiment Team}.
\newblock Open {X}-embodiment: Robotic learning datasets and {RT-X} models : Open {X}-embodiment collaboration.
\newblock In \emph{IEEE International Conference on Robotics and Automation}, 2024.

\bibitem[Pathak et~al.(2019)Pathak, Lu, Darrell, Isola, and Efros]{PathakLDIE19SelfAssemb}
Deepak Pathak, Christopher Lu, Trevor Darrell, Phillip Isola, and Alexei~A. Efros.
\newblock Learning to control self-assembling morphologies: {A} study of generalization via modularity.
\newblock In \emph{Advances in Neural Information Processing Systems}, 2019.

\bibitem[Reed et~al.(2022)Reed, Zolna, Parisotto, Colmenarejo, Novikov, Barth-Maron, Gimenez, Sulsky, Kay, Springenberg, et~al.]{reed2022generalist}
Scott Reed, Konrad Zolna, Emilio Parisotto, Sergio~Gomez Colmenarejo, Alexander Novikov, Gabriel Barth-Maron, Mai Gimenez, Yury Sulsky, Jackie Kay, Jost~Tobias Springenberg, et~al.
\newblock A generalist agent.
\newblock \emph{CoRR}, abs/2205.06175, 2022.

\bibitem[Scarselli et~al.(2009)Scarselli, Gori, Tsoi, Hagenbuchner, and Monfardini]{ScarselliGTHM09GNN}
Franco Scarselli, Marco Gori, Ah~Chung Tsoi, Markus Hagenbuchner, and Gabriele Monfardini.
\newblock The graph neural network model.
\newblock \emph{IEEE Transactions on Neural Networks}, 20\penalty0 (1):\penalty0 61--80, 2009.

\bibitem[Schaff et~al.(2019)Schaff, Yunis, Chakrabarti, and Walter]{SchaffYCW19Jointly}
Charles~B. Schaff, David Yunis, Ayan Chakrabarti, and Matthew~R. Walter.
\newblock Jointly learning to construct and control agents using deep reinforcement learning.
\newblock In \emph{International Conference on Robotics and Automation}, 2019.

\bibitem[Schulman et~al.(2017)Schulman, Wolski, Dhariwal, Radford, and Klimov]{schulman2017PPO}
John Schulman, Filip Wolski, Prafulla Dhariwal, Alec Radford, and Oleg Klimov.
\newblock Proximal policy optimization algorithms.
\newblock \emph{CoRR}, abs/1707.06347, 2017.

\bibitem[Sferrazza et~al.(2024)Sferrazza, Huang, Liu, Lee, and Abbeel]{sferrazza2024bodytransformer}
Carmelo Sferrazza, Dun-Ming Huang, Fangchen Liu, Jongmin Lee, and Pieter Abbeel.
\newblock Body transformer: Leveraging robot embodiment for policy learning.
\newblock In \emph{Workshop on Embodiment-Aware Robot Learning}, 2024.

\bibitem[Trabucco et~al.(2022)Trabucco, Phielipp, and Berseth]{trabucco2022anymorph}
Brandon Trabucco, Mariano Phielipp, and Glen Berseth.
\newblock {A}ny{M}orph: Learning transferable polices by inferring agent morphology.
\newblock In \emph{International Conference on Machine Learning}, 2022.

\bibitem[Tsai et~al.(2020)Tsai, Chen, and Ho]{tsai2020transferlearningwithoutknowing}
Yun-Yun Tsai, Pin-Yu Chen, and Tsung-Yi Ho.
\newblock Transfer learning without knowing: Reprogramming black-box machine learning models with scarce data and limited resources.
\newblock In \emph{International Conference on Machine Learning}, 2020.

\bibitem[Vaswani et~al.(2017)Vaswani, Shazeer, Parmar, Uszkoreit, Jones, Gomez, Kaiser, and Polosukhin]{VaswaniSPUJGKP17Attention}
Ashish Vaswani, Noam Shazeer, Niki Parmar, Jakob Uszkoreit, Llion Jones, Aidan~N. Gomez, Lukasz Kaiser, and Illia Polosukhin.
\newblock Attention is all you need.
\newblock In \emph{Advances in Neural Information Processing Systems}, 2017.

\bibitem[Wang et~al.(2018)Wang, Liao, Ba, and Fidler]{wang2018nervenet}
Tingwu Wang, Renjie Liao, Jimmy Ba, and Sanja Fidler.
\newblock {NerveNet}: Learning structured policy with graph neural networks.
\newblock In \emph{International Conference on Learning Representations}, 2018.

\bibitem[Wang et~al.(2022)Wang, Zhang, Lee, Zhang, Sun, Ren, Su, Perot, Dy, and Pfister]{Wang2022promptingInContinualLearning}
Zifeng Wang, Zizhao Zhang, Chen-Yu Lee, Han Zhang, Ruoxi Sun, Xiaoqi Ren, Guolong Su, Vincent Perot, Jennifer Dy, and Tomas Pfister.
\newblock Learning to prompt for continual learning.
\newblock In \emph{IEEE/CVF Conference on Computer Vision and Pattern Recognition}, 2022.

\bibitem[Xiong et~al.(2023)Xiong, Beck, and Whiteson]{xiong2023modumorph}
Zheng Xiong, Jacob Beck, and Shimon Whiteson.
\newblock Universal morphology control via contextual modulation.
\newblock In \emph{International Conference on Machine Learning}, 2023.

\bibitem[Xiong et~al.(2024)Xiong, Vuorio, Beck, Zimmer, Shao, and Whiteson]{xiong2024distillingHypernetworks}
Zheng Xiong, Risto Vuorio, Jacob Beck, Matthieu Zimmer, Kun Shao, and Shimon Whiteson.
\newblock Distilling morphology-conditioned hypernetworks for efficient universal morphology control.
\newblock \emph{CoRR}, abs/2402.06570, 2024.

\bibitem[Yuan et~al.(2022)Yuan, Song, Luo, Sun, and Kitani]{Ye22Transform2Act}
Ye~Yuan, Yuda Song, Zhengyi Luo, Wen Sun, and Kris~M. Kitani.
\newblock Transform2act: Learning a transform-and-control policy for efficient agent design.
\newblock In \emph{International Conference on Learning Representations}, 2022.

\bibitem[Zheng et~al.(2023)Zheng, Feng, Xia, Jiang, Demontis, Pintor, Biggio, and Roli]{yang2023WhyAdvReprograWorksWhenfail}
Yang Zheng, Xiaoyi Feng, Zhaoqiang Xia, Xiaoyue Jiang, Ambra Demontis, Maura Pintor, Battista Biggio, and Fabio Roli.
\newblock Why adversarial reprogramming works, when it fails, and how to tell the difference.
\newblock \emph{Information Sciences}, 632:\penalty0 130--143, 2023.

\end{thebibliography}
\bibliographystyle{rlj}

\beginSupplementaryMaterials

\section{Direct Finetuning Configurations}
\label{sec:appendix_directfinetuning}

In our experiments, we consider various finetuning scenarios in our evaluations. For direct finetuning methods, we include combinations of subsets we finetune online in Table~\ref{tab:layer_tuning}. Our evaluations included subsets of the direct tuning configurations of weight combinations. For example, \textit{Input Embedding} includes combinations in which just $W^{\text{embed}}$, $W^{\text{position}}$ and both $\{W^{\text{emebed}}, W^{\text{position}}\}$ are tuned online during training.

\begin{table}
\hspace{10mm}
\centering
\caption{Layer tuning parameters and experiment identifiers}
\label{tab:layer_tuning}
\begin{tabular}{lll}
\toprule
\textbf{Layer Tuned} & \textbf{Parameters $\phi$} & \textbf{Exp. Identifier} \\
\midrule
End-to-end & $\theta^*$ & E2E \\
Transformer layers & $\{T_i; i \in [1, L]\}$ & Layer 5 \\
Attention layers & $\{W^{\text{attn}}_i; i \in [1, L]\}$ & Lora \\
Nonlinear transformation & $\{W^{\text{in}}_i, W^{\text{out}}_i; i \in [1, L]\}$ & Lora \\
Input Embedding & $\{W^{\text{embed}}, W^{\text{position}}\}$ & Embedding \\
Decoder & $\{W^{\text{decoder}}_i ; i \in [1, L^{\text{dec}}]\}$ & Decoder \\
\bottomrule
\end{tabular}
\end{table}

\section{LoRA Initialization Details}
\label{sec:appendix_lora}
When using LoRA in our experiments, we initialize $B$ to small Gaussian noise $b_{ij} \sim N(0; 10^{-4})$ and A to a zero matrix which eliminates LoRA adapters affect on the zeros-hot performance at the beginning of training. LoRA was included as a  finetuning method because we want to reduce the total number of parameters used which LoRA can explicitly do via the rank.

\section{Morphology-Aware Policy Performance }
\label{sec:appendix_best PEFT}

This section reports results for the best-performing PEFT algorithms for each significant grouping of methods we consider. Table~\ref{tab:group_performance_ft} show flat terrain results, Table~\ref{tab:group_performance_csr} shows variable terrain results, and Table~\ref{tab:group_performance_obstacle} shows results for obstacle avoidance. These results report statistical significance when comparing results to zero-shot pretraining performance and training policies from scratch. Surprisingly, training from scratch worked surprisingly well in flat terrain. Still, most PEFT techniques perform better after five million samples than training from scratch on more complex tasks. 

\begin{table}[h]
\centering
\caption{Flat Terrain Cumulative Rewards for each testing morphology.  Values show mean (top) and standard deviation (bottom). $^\dagger$ statistical significance compared to Zero Shot and $\ddagger$ statistical significance to Scratch (p < 0.01). P-values and the hypothesis test run (T: t-test, M: Mann-Whitney) comparing against Zero shot and Scratch results.}

{\small
\begin{tabular}{l@{\hspace{3pt}}c@{\hspace{3pt}}c@{\hspace{3pt}}c@{\hspace{3pt}}c@{\hspace{3pt}}c@{\hspace{3pt}}c}
\hline
Morphology & 1 & 2 & 3 & 4 & 5 & 6 \\
\hline
Full Model & $4281.46$$^\dagger$$^\ddagger$ & $4552.77$$^\dagger$$^\ddagger$ & $1635.82$$^\dagger$ & $5545.44$$^\dagger$$^\ddagger$ & $5019.71$$^\dagger$$^\ddagger$ & $5558.61$$^\dagger$$^\ddagger$  \\
 & $\pm 181.77$ & $\pm 239.11$ & $\pm 405.64$ & $\pm 280.16$ & $\pm 143.13$ & $\pm 286.80$  \\
Layer 4 & $3761.11$$^\dagger$ & $4121.84$$^\dagger$ & $1491.06$$^\dagger$ & $5183.88$$^\dagger$$^\ddagger$ & $4666.95$$^\dagger$$^\ddagger$ & $5192.58$$^\dagger$$^\ddagger$  \\
 & $\pm 102.11$ & $\pm 120.48$ & $\pm 230.10$ & $\pm 167.91$ & $\pm 255.22$ & $\pm 152.33$  \\
Lora & $3798.90$$^\dagger$ & $4208.41$$^\dagger$$^\ddagger$ & $1639.69$$^\dagger$ & $5223.47$$^\dagger$$^\ddagger$ & $4761.58$$^\dagger$$^\ddagger$ & $5223.47$$^\dagger$$^\ddagger$  \\
 & $\pm 138.55$ & $\pm 77.12$ & $\pm 41.31$ & $\pm 174.25$ & $\pm 210.57$ & $\pm 174.25$  \\
Decoder Only & $2732.26$$^\dagger$$^\ddagger$ & $3112.46$$^\dagger$$^\ddagger$ & $1398.42$$^\dagger$ & $4868.54$$^\ddagger$ & $3404.67$$^\dagger$$^\ddagger$ & $4858.76$$^\ddagger$  \\
 & $\pm 71.35$ & $\pm 221.65$ & $\pm 210.79$ & $\pm 263.81$ & $\pm 92.07$ & $\pm 248.01$  \\
Embeding & $3308.43$$^\dagger$$^\ddagger$ & $3684.66$$^\dagger$ & $1554.28$$^\dagger$ & $4986.16$$^\ddagger$ & $4062.05$$^\dagger$ & $4997.82$$^\ddagger$  \\
 & $\pm 115.83$ & $\pm 104.99$ & $\pm 190.82$ & $\pm 183.78$ & $\pm 248.20$ & $\pm 191.07$  \\
Input Adapt & $3231.84$$^\dagger$$^\ddagger$ & $3529.41$$^\dagger$ & $1510.46$$^\dagger$ & $4927.72$$^\ddagger$ & $3946.59$$^\dagger$ & $4963.53$$^\ddagger$  \\
 & $\pm 100.03$ & $\pm 104.58$ & $\pm 242.36$ & $\pm 225.75$ & $\pm 220.78$ & $\pm 222.35$  \\
Prefix & $3332.33$$^\dagger$$^\ddagger$ & $3750.54$$^\dagger$ & $1604.92$$^\dagger$ & $5064.15$$^\ddagger$ & $4199.89$$^\dagger$ & $5066.47$$^\ddagger$  \\
 & $\pm 126.28$ & $\pm 201.62$ & $\pm 336.88$ & $\pm 133.91$ & $\pm 276.27$ & $\pm 137.63$  \\
Scratch & $3754.15$$^\dagger$ & $3840.33$$^\dagger$ & $2191.50$$^\dagger$ & $3727.29$ & $4085.82$$^\dagger$ & $3608.55$  \\
 & $\pm 210.65$ & $\pm 211.59$ & $\pm 624.72$ & $\pm 733.60$ & $\pm 217.00$ & $\pm 777.33$  \\
Zero Shot & $1867.58$ & $1703.19$ & $253.70$ & $4392.08$ & $1849.41$ & $4431.93$  \\
 & $\pm 82.55$ & $\pm 447.69$ & $\pm 188.25$ & $\pm 434.01$ & $\pm 338.10$ & $\pm 405.78$  \\
\hline
 & \multicolumn{6}{c}{P-Values comparing against Zeroshot Performance$^\dagger$}  \\
\hline
Full Model & $9.1 \times 10^{-9}({T} )$ & $3.6 \times 10^{-6}({T} )$ & $2.6 \times 10^{-4}({T} )$ & $2.1 \times 10^{-3}({T} )$ & $1.3 \times 10^{-7}({T} )$ & $1.9 \times 10^{-3}({T} )$ \\
Layer 4 & $2.3 \times 10^{-9}({T} )$ & $6.2 \times 10^{-6}({T} )$ & $3.3 \times 10^{-5}({T} )$ & $9.3 \times 10^{-3}({T} )$ & $9.7 \times 10^{-7}({T} )$ & $8.0 \times 10^{-3}({T} )$ \\
Lora & $9.8 \times 10^{-9}({T} )$ & $4.1 \times 10^{-6}({T} )$ & $5.3 \times 10^{-7}({T} )$ & $7.5 \times 10^{-3}({T} )$ & $4.7 \times 10^{-7}({T} )$ & $7.1 \times 10^{-3}({T} )$ \\
Decoder Only & $2.5 \times 10^{-7}({T} )$ & $4.9 \times 10^{-4}({T} )$ & $4.0 \times 10^{-5}({T} )$ & $9.7 \times 10^{-2}({T} )$ & $2.1 \times 10^{-5}({T} )$ & $1.1 \times 10^{-1}({T} )$ \\
Embeding & $3.7 \times 10^{-8}({T} )$ & $2.5 \times 10^{-5}({T} )$ & $7.9 \times 10^{-3}({M} )$ & $3.6 \times 10^{-2}({T} )$ & $5.7 \times 10^{-6}({T} )$ & $3.6 \times 10^{-2}({T} )$ \\
Input Adapt & $2.7 \times 10^{-8}({T} )$ & $4.6 \times 10^{-5}({T} )$ & $7.9 \times 10^{-3}({M} )$ & $6.0 \times 10^{-2}({T} )$ & $6.4 \times 10^{-6}({T} )$ & $5.1 \times 10^{-2}({T} )$ \\
Prefix & $5.1 \times 10^{-8}({T} )$ & $3.2 \times 10^{-5}({T} )$ & $1.1 \times 10^{-4}({T} )$ & $1.8 \times 10^{-2}({T} )$ & $4.9 \times 10^{-6}({T} )$ & $1.8 \times 10^{-2}({T} )$ \\
Scratch & $0.0({T} )$ & $2.9 \times 10^{-8}({T} )$ & $2.7 \times 10^{-5}({T} )$ & $1.1 \times 10^{-1}({T} )$ & $2.2 \times 10^{-9}({T} )$ & $5.9 \times 10^{-2}({T} )$ \\
Zero Shot & $1.0({T} )$ & $1.0({T} )$ & $1.0({T} )$ & $1.0({T} )$ & $1.0({T} )$ & $1.0({T} )$ \\
\hline
 & \multicolumn{6}{c}{P-Values comparing against Scratch Performance$^\ddagger$}  \\
\hline
Full Model & $6.6 \times 10^{-4}({T} )$ & $1.1 \times 10^{-4}({T} )$ & $1.2 \times 10^{-1}({T} )$ & $2.5 \times 10^{-4}({T} )$ & $1.9 \times 10^{-6}({T} )$ & $2.2 \times 10^{-4}({T} )$ \\
Layer 4 & $9.5 \times 10^{-1}({T} )$ & $2.3 \times 10^{-2}({T} )$ & $4.2 \times 10^{-2}({T} )$ & $1.3 \times 10^{-3}({T} )$ & $8.9 \times 10^{-4}({T} )$ & $1.0 \times 10^{-3}({T} )$ \\
Lora & $6.9 \times 10^{-1}({T} )$ & $3.9 \times 10^{-3}({T} )$ & $8.9 \times 10^{-2}({T} )$ & $1.1 \times 10^{-3}({T} )$ & $1.3 \times 10^{-4}({T} )$ & $9.1 \times 10^{-4}({T} )$ \\
Decoder Only & $2.2 \times 10^{-7}({T} )$ & $6.7 \times 10^{-5}({T} )$ & $2.3 \times 10^{-2}({T} )$ & $7.8 \times 10^{-3}({T} )$ & $2.9 \times 10^{-5}({T} )$ & $6.1 \times 10^{-3}({T} )$ \\
Embeding & $1.2 \times 10^{-3}({T} )$ & $1.7 \times 10^{-1}({T} )$ & $1.6 \times 10^{-1}({M} )$ & $3.8 \times 10^{-3}({T} )$ & $8.6 \times 10^{-1}({T} )$ & $2.9 \times 10^{-3}({T} )$ \\
Input Adapt & $2.9 \times 10^{-4}({T} )$ & $1.3 \times 10^{-2}({T} )$ & $9.9 \times 10^{-2}({M} )$ & $5.4 \times 10^{-3}({T} )$ & $3.0 \times 10^{-1}({T} )$ & $3.5 \times 10^{-3}({T} )$ \\
Prefix & $2.1 \times 10^{-3}({T} )$ & $4.8 \times 10^{-1}({T} )$ & $9.1 \times 10^{-2}({T} )$ & $2.4 \times 10^{-3}({T} )$ & $4.3 \times 10^{-1}({T} )$ & $1.9 \times 10^{-3}({T} )$ \\
Scratch & $1.0({T} )$ & $1.0({T} )$ & $1.0({T} )$ & $1.0({T} )$ & $1.0({T} )$ & $1.0({T} )$ \\
Zero Shot & $0.0({T} )$ & $2.9 \times 10^{-8}({T} )$ & $2.7 \times 10^{-5}({T} )$ & $1.1 \times 10^{-1}({T} )$ & $2.2 \times 10^{-9}({T} )$ & $5.9 \times 10^{-2}({T} )$ \\
\hline
\end{tabular}
}
\label{tab:group_performance_ft}
\end{table}

\begin{table}[h]
\centering
\caption{Variable Terrain Cumulative Rewards for each testing morphology.  Values show mean (top) and standard deviation (bottom). $^\dagger$ statistical significance compared to Zero Shot and $\ddagger$ statistical significance to Scratch (p < 0.01). P-values and the hypothesis test run (T: t-test, M: Mann-Whitney) comparing against Zero shot and Scratch results.}
{\small
\begin{tabular}{l@{\hspace{3pt}}c@{\hspace{3pt}}c@{\hspace{3pt}}c@{\hspace{3pt}}c@{\hspace{3pt}}c@{\hspace{3pt}}c}
\hline
Morphology & 1 & 2 & 3 & 4 & 5 & 6 \\
\hline
Full Model & $2253.96$$^\dagger$$^\ddagger$ & $1983.81$$^\dagger$$^\ddagger$ & $2001.18$$^\dagger$$^\ddagger$ & $3560.43$$^\dagger$$^\ddagger$ & $2047.49$$^\dagger$$^\ddagger$ & $3595.38$$^\dagger$$^\ddagger$  \\
 & $\pm 41.47$ & $\pm 154.82$ & $\pm 42.14$ & $\pm 317.89$ & $\pm 117.06$ & $\pm 368.99$  \\
Layer 4 & $2093.75$$^\dagger$$^\ddagger$ & $1871.09$$^\dagger$$^\ddagger$ & $1879.22$$^\dagger$$^\ddagger$ & $3254.06$$^\dagger$$^\ddagger$ & $1912.17$$^\dagger$$^\ddagger$ & $3279.03$$^\ddagger$  \\
 & $\pm 34.23$ & $\pm 79.86$ & $\pm 33.91$ & $\pm 353.70$ & $\pm 135.29$ & $\pm 379.46$  \\
Lora & $2141.39$$^\dagger$$^\ddagger$ & $1848.53$$^\dagger$$^\ddagger$ & $1786.88$$^\dagger$$^\ddagger$ & $3230.13$$^\dagger$$^\ddagger$ & $1878.93$$^\dagger$$^\ddagger$ & $3234.25$$^\ddagger$  \\
 & $\pm 53.29$ & $\pm 113.44$ & $\pm 72.97$ & $\pm 327.39$ & $\pm 107.89$ & $\pm 329.42$  \\
Decoder Only & $1969.63$$^\dagger$$^\ddagger$ & $1623.70$$^\dagger$$^\ddagger$ & $1299.89$$^\dagger$ & $3164.72$$^\dagger$$^\ddagger$ & $1672.47$$^\dagger$$^\ddagger$ & $3180.90$$^\ddagger$  \\
 & $\pm 28.01$ & $\pm 126.14$ & $\pm 70.71$ & $\pm 307.28$ & $\pm 112.14$ & $\pm 316.43$  \\
Embeding & $1836.54$$^\dagger$$^\ddagger$ & $1529.38$$^\dagger$ & $1441.65$$^\dagger$ & $2872.51$$^\ddagger$ & $1549.29$$^\dagger$ & $2887.67$$^\ddagger$  \\
 & $\pm 25.22$ & $\pm 84.38$ & $\pm 41.51$ & $\pm 307.30$ & $\pm 106.71$ & $\pm 311.40$  \\
Input Adapt & $1820.01$$^\dagger$$^\ddagger$ & $1521.18$$^\dagger$ & $1338.57$$^\dagger$ & $2869.53$$^\ddagger$ & $1512.25$$^\dagger$ & $2895.01$$^\ddagger$  \\
 & $\pm 48.63$ & $\pm 106.76$ & $\pm 61.81$ & $\pm 293.57$ & $\pm 109.46$ & $\pm 299.56$  \\
Prefix & $1902.95$$^\dagger$$^\ddagger$ & $1643.33$$^\dagger$$^\ddagger$ & $1406.55$$^\dagger$ & $2930.13$$^\ddagger$ & $1601.95$$^\dagger$ & $2918.47$$^\ddagger$  \\
 & $\pm 43.36$ & $\pm 165.26$ & $\pm 83.55$ & $\pm 261.58$ & $\pm 134.90$ & $\pm 300.10$  \\
Scratch & $1679.33$$^\dagger$ & $1406.59$$^\dagger$ & $1406.58$$^\dagger$ & $1735.22$$^\dagger$ & $1449.59$$^\dagger$ & $1758.99$$^\dagger$  \\
 & $\pm 82.91$ & $\pm 101.71$ & $\pm 164.55$ & $\pm 166.72$ & $\pm 69.66$ & $\pm 168.21$  \\
Zero Shot & $1259.92$ & $591.83$ & $136.82$ & $2452.59$ & $685.54$ & $2476.77$  \\
 & $\pm 61.93$ & $\pm 67.70$ & $\pm 103.66$ & $\pm 291.96$ & $\pm 71.67$ & $\pm 349.00$  \\
\hline
& \multicolumn{6}{c}{P-Values comparing against Zeroshot Performance$^\dagger$}  \\
\hline
Full Model & $4.6 \times 10^{-9}({T} )$ & $2.1 \times 10^{-6}({T} )$ & $5.2 \times 10^{-6}({T} )$ & $4.7 \times 10^{-9}({T} )$ & $3.5 \times 10^{-8}({T} )$ & $1.5 \times 10^{-8}({T} )$ \\
Layer 4 & $1.8 \times 10^{-7}({T} )$ & $1.5 \times 10^{-6}({T} )$ & $5.1 \times 10^{-5}({T} )$ & $6.7 \times 10^{-4}({M} )$ & $1.9 \times 10^{-6}({T} )$ & $6.7 \times 10^{-4}({M} )$ \\
Lora & $9.5 \times 10^{-8}({T} )$ & $8.0 \times 10^{-6}({T} )$ & $5.1 \times 10^{-4}({T} )$ & $6.7 \times 10^{-4}({M} )$ & $9.4 \times 10^{-7}({T} )$ & $6.7 \times 10^{-4}({M} )$ \\
Decoder Only & $8.2 \times 10^{-6}({T} )$ & $5.3 \times 10^{-3}({T} )$ & $2.2 \times 10^{-1}({T} )$ & $6.7 \times 10^{-4}({M} )$ & $7.2 \times 10^{-4}({T} )$ & $6.7 \times 10^{-4}({M} )$ \\
Embeding & $2.0 \times 10^{-3}({T} )$ & $4.9 \times 10^{-2}({T} )$ & $6.7 \times 10^{-1}({T} )$ & $6.7 \times 10^{-4}({M} )$ & $6.4 \times 10^{-2}({T} )$ & $6.7 \times 10^{-4}({M} )$ \\
Input Adapt & $6.2 \times 10^{-3}({T} )$ & $8.2 \times 10^{-2}({T} )$ & $4.2 \times 10^{-1}({T} )$ & $6.7 \times 10^{-4}({M} )$ & $2.3 \times 10^{-1}({T} )$ & $6.7 \times 10^{-4}({M} )$ \\
Prefix & $4.8 \times 10^{-4}({T} )$ & $7.2 \times 10^{-3}({T} )$ & $1.0 \times 10^{0}({T} )$ & $6.7 \times 10^{-4}({M} )$ & $1.9 \times 10^{-2}({T} )$ & $6.7 \times 10^{-4}({M} )$ \\
Scratch & $1.0({T} )$ & $1.0({T} )$ & $1.0({T} )$ & $1.0({T} )$ & $1.0({T} )$ & $1.0({T} )$ \\
Zero Shot & $4.1 \times 10^{-7}({T} )$ & $1.3 \times 10^{-9}({T} )$ & $1.8 \times 10^{-9}({T} )$ & $2.7 \times 10^{-3}({M} )$ & $0.0({T} )$ & $2.4 \times 10^{-4}({T} )$ \\
\hline
 & \multicolumn{6}{c}{P-Values comparing against Scratch Performance$^\ddagger$}  \\
\hline
Full Model & $1.5 \times 10^{-3}({T} )$ & $3.2 \times 10^{-8}({T} )$ & $1.3 \times 10^{-2}({T} )$ & $3.7 \times 10^{-7}({T} )$ & $9.3 \times 10^{-8}({T} )$ & $1.3 \times 10^{-7}({T} )$ \\
Layer 4 & $2.7 \times 10^{-1}({T} )$ & $1.1 \times 10^{-6}({T} )$ & $2.6 \times 10^{-3}({T} )$ & $2.9 \times 10^{-6}({T} )$ & $2.3 \times 10^{-7}({T} )$ & $2.7 \times 10^{-6}({T} )$ \\
Lora & $5.6 \times 10^{-3}({T} )$ & $1.5 \times 10^{-4}({T} )$ & $3.6 \times 10^{-3}({T} )$ & $7.3 \times 10^{-6}({T} )$ & $1.3 \times 10^{-6}({T} )$ & $1.2 \times 10^{-6}({T} )$ \\
Decoder Only & $3.8 \times 10^{-1}({T} )$ & $2.1 \times 10^{-4}({T} )$ & $7.5 \times 10^{-4}({T} )$ & $5.2 \times 10^{-5}({T} )$ & $2.3 \times 10^{-4}({T} )$ & $1.5 \times 10^{-6}({T} )$ \\
Embeding & $9.8 \times 10^{-7}({T} )$ & $4.1 \times 10^{-1}({T} )$ & $6.7 \times 10^{-4}({M} )$ & $3.6 \times 10^{-4}({T} )$ & $1.0 \times 10^{-1}({T} )$ & $3.7 \times 10^{-4}({T} )$ \\
Input Adapt & $1.5 \times 10^{-6}({T} )$ & $5.9 \times 10^{-1}({T} )$ & $5.7 \times 10^{-4}({T} )$ & $2.0 \times 10^{-4}({T} )$ & $2.1 \times 10^{-1}({T} )$ & $7.7 \times 10^{-5}({T} )$ \\
Prefix & $1.3 \times 10^{-5}({T} )$ & $1.3 \times 10^{-2}({T} )$ & $1.2 \times 10^{-3}({T} )$ & $2.2 \times 10^{-4}({T} )$ & $2.3 \times 10^{-3}({T} )$ & $1.2 \times 10^{-5}({T} )$ \\
Scratch & $1.0({T} )$ & $1.0({T} )$ & $1.0({T} )$ & $1.0({T} )$ & $1.0({T} )$ & $1.0({T} )$ \\
Zero Shot & $0.0({T} )$ & $5.6 \times 10^{-7}({T} )$ & $0.0({T} )$ & $2.0 \times 10^{-2}({T} )$ & $8.8 \times 10^{-8}({T} )$ & $8.3 \times 10^{-3}({T} )$ \\
\hline

\end{tabular}
}
\label{tab:group_performance_csr}
\end{table}

\begin{table}[h]
\centering
\caption{Obstacle Avoidance Cumulative Rewards for each testing morphology.  Values show mean (top) and standard deviation (bottom). $^\dagger$ statistical significance compared to Zero Shot and $\ddagger$ statistical significance to Scratch (p < 0.01). P-values and the hypothesis test run (T: t-test, M: Mann-Whitney) comparing against Zero shot and Scratch results.}
{\small
\begin{tabular}{l@{\hspace{3pt}}c@{\hspace{3pt}}c@{\hspace{3pt}}c@{\hspace{3pt}}c@{\hspace{3pt}}c@{\hspace{3pt}}c}
\hline
Morphology & 1 & 2 & 3 & 4 & 5 & 6 \\
\hline
Full Model & $2652.41$$^\dagger$$^\ddagger$ & $3101.42$$^\dagger$$^\ddagger$ & $1705.64$$^\dagger$ & $3577.09$$^\dagger$$^\ddagger$ & $3219.75$$^\dagger$$^\ddagger$ & $3558.26$$^\dagger$$^\ddagger$  \\
 & $\pm 193.57$ & $\pm 177.17$ & $\pm 4.16$ & $\pm 341.74$ & $\pm 199.13$ & $\pm 365.21$  \\
Layer 4 & $2246.88$$^\dagger$ & $2684.70$$^\dagger$$^\ddagger$ & $1592.29$$^\dagger$$^\ddagger$ & $3276.76$$^\dagger$$^\ddagger$ & $2888.34$$^\dagger$$^\ddagger$ & $3194.19$$^\ddagger$  \\
 & $\pm 184.03$ & $\pm 85.63$ & $\pm 140.05$ & $\pm 314.17$ & $\pm 64.04$ & $\pm 351.87$  \\
Lora & $2137.75$$^\dagger$$^\ddagger$ & $2585.71$$^\dagger$$^\ddagger$ & $1672.75$$^\dagger$$^\ddagger$ & $3191.40$$^\dagger$$^\ddagger$ & $2851.48$$^\dagger$$^\ddagger$ & $3189.01$$^\ddagger$  \\
 & $\pm 116.21$ & $\pm 191.00$ & $\pm 12.63$ & $\pm 320.51$ & $\pm 107.16$ & $\pm 319.76$  \\
Decoder Only & $2263.74$$^\dagger$ & $2531.13$$^\dagger$$^\ddagger$ & $1456.02$$^\dagger$$^\ddagger$ & $3061.26$$^\ddagger$ & $2672.06$$^\dagger$$^\ddagger$ & $3132.18$$^\ddagger$  \\
 & $\pm 186.99$ & $\pm 161.72$ & $\pm 218.88$ & $\pm 360.37$ & $\pm 160.55$ & $\pm 302.67$  \\
Embeding & $1863.25$$^\dagger$$^\ddagger$ & $2189.46$$^\dagger$ & $1556.72$$^\dagger$$^\ddagger$ & $2882.24$$^\ddagger$ & $2398.29$$^\dagger$ & $2877.35$$^\ddagger$  \\
 & $\pm 94.00$ & $\pm 167.27$ & $\pm 151.65$ & $\pm 361.86$ & $\pm 139.50$ & $\pm 417.09$  \\
Input Adapt & $1839.40$$^\dagger$$^\ddagger$ & $2159.73$$^\dagger$ & $1458.49$$^\dagger$$^\ddagger$ & $2929.91$$^\ddagger$ & $2367.39$$^\dagger$ & $2833.04$$^\ddagger$  \\
 & $\pm 117.50$ & $\pm 125.84$ & $\pm 206.98$ & $\pm 356.45$ & $\pm 142.90$ & $\pm 312.07$  \\
Prefix & $1841.45$$^\dagger$$^\ddagger$ & $2334.01$$^\dagger$ & $1514.42$$^\dagger$$^\ddagger$ & $2877.43$$^\ddagger$ & $2538.31$$^\dagger$$^\ddagger$ & $2935.63$$^\ddagger$  \\
 & $\pm 142.33$ & $\pm 133.00$ & $\pm 186.34$ & $\pm 324.42$ & $\pm 119.05$ & $\pm 293.07$  \\
Scratch & $2334.75$$^\dagger$ & $2119.16$$^\dagger$ & $1843.23$$^\dagger$ & $2112.34$ & $2265.47$$^\dagger$ & $2144.60$$^\dagger$  \\
 & $\pm 92.45$ & $\pm 124.11$ & $\pm 99.74$ & $\pm 216.38$ & $\pm 124.23$ & $\pm 123.50$  \\
Zero Shot & $1300.21$ & $1184.87$ & $332.64$ & $2467.45$ & $1295.92$ & $2488.64$  \\
 & $\pm 117.01$ & $\pm 248.07$ & $\pm 114.42$ & $\pm 246.89$ & $\pm 202.99$ & $\pm 274.84$  \\
\hline
& \multicolumn{6}{c}{P-Values comparing against Zeroshot Performance$^\dagger$}  \\
\hline
Full Model & $2.2 \times 10^{-6}({T} )$ & $1.5 \times 10^{-6}({T} )$ & $9.7 \times 10^{-9}({T} )$ & $7.6 \times 10^{-4}({T} )$ & $8.5 \times 10^{-7}({T} )$ & $1.6 \times 10^{-3}({T} )$ \\
Layer 4 & $2.4 \times 10^{-5}({T} )$ & $3.1 \times 10^{-6}({T} )$ & $6.8 \times 10^{-7}({T} )$ & $3.7 \times 10^{-3}({T} )$ & $3.9 \times 10^{-7}({T} )$ & $1.3 \times 10^{-2}({T} )$ \\
Lora & $7.6 \times 10^{-6}({T} )$ & $1.9 \times 10^{-5}({T} )$ & $1.2 \times 10^{-8}({T} )$ & $7.2 \times 10^{-3}({T} )$ & $8.4 \times 10^{-7}({T} )$ & $1.1 \times 10^{-2}({T} )$ \\
Decoder Only & $2.3 \times 10^{-5}({T} )$ & $1.7 \times 10^{-5}({T} )$ & $1.7 \times 10^{-5}({T} )$ & $2.6 \times 10^{-2}({T} )$ & $5.4 \times 10^{-6}({T} )$ & $1.4 \times 10^{-2}({T} )$ \\
Embeding & $6.9 \times 10^{-5}({T} )$ & $1.5 \times 10^{-4}({T} )$ & $7.9 \times 10^{-3}({M} )$ & $9.5 \times 10^{-2}({T} )$ & $1.9 \times 10^{-5}({T} )$ & $1.6 \times 10^{-1}({T} )$ \\
Input Adapt & $1.9 \times 10^{-4}({T} )$ & $1.1 \times 10^{-4}({T} )$ & $1.2 \times 10^{-5}({T} )$ & $6.5 \times 10^{-2}({T} )$ & $2.5 \times 10^{-5}({T} )$ & $1.4 \times 10^{-1}({T} )$ \\
Prefix & $8.9 \times 10^{-4}({T} )$ & $3.8 \times 10^{-5}({T} )$ & $4.7 \times 10^{-6}({T} )$ & $7.9 \times 10^{-2}({T} )$ & $5.6 \times 10^{-6}({T} )$ & $5.7 \times 10^{-2}({T} )$ \\
Scratch & $0.0({T} )$ & $5.6 \times 10^{-7}({T} )$ & $0.0({T} )$ & $2.0 \times 10^{-2}({T} )$ & $8.8 \times 10^{-8}({T} )$ & $8.3 \times 10^{-3}({T} )$ \\
Zero Shot & $1.0({T} )$ & $1.0({T} )$ & $1.0({T} )$ & $1.0({T} )$ & $1.0({T} )$ & $1.0({T} )$ \\
\hline
 & \multicolumn{6}{c}{P-Values comparing against Scratch Performance$^\ddagger$}  \\
\hline
Full Model & $1.5 \times 10^{-3}({T} )$ & $3.2 \times 10^{-8}({T} )$ & $1.3 \times 10^{-2}({T} )$ & $3.7 \times 10^{-7}({T} )$ & $9.3 \times 10^{-8}({T} )$ & $1.3 \times 10^{-7}({T} )$ \\
Layer 4 & $2.7 \times 10^{-1}({T} )$ & $1.1 \times 10^{-6}({T} )$ & $2.6 \times 10^{-3}({T} )$ & $2.9 \times 10^{-6}({T} )$ & $2.3 \times 10^{-7}({T} )$ & $2.7 \times 10^{-6}({T} )$ \\
Lora & $5.6 \times 10^{-3}({T} )$ & $1.5 \times 10^{-4}({T} )$ & $3.6 \times 10^{-3}({T} )$ & $7.3 \times 10^{-6}({T} )$ & $1.3 \times 10^{-6}({T} )$ & $1.2 \times 10^{-6}({T} )$ \\
Decoder Only & $3.8 \times 10^{-1}({T} )$ & $2.1 \times 10^{-4}({T} )$ & $7.5 \times 10^{-4}({T} )$ & $5.2 \times 10^{-5}({T} )$ & $2.3 \times 10^{-4}({T} )$ & $1.5 \times 10^{-6}({T} )$ \\
Embeding & $9.8 \times 10^{-7}({T} )$ & $4.1 \times 10^{-1}({T} )$ & $6.7 \times 10^{-4}({M} )$ & $3.6 \times 10^{-4}({T} )$ & $1.0 \times 10^{-1}({T} )$ & $3.7 \times 10^{-4}({T} )$ \\
Input Adapt & $1.5 \times 10^{-6}({T} )$ & $5.9 \times 10^{-1}({T} )$ & $5.7 \times 10^{-4}({T} )$ & $2.0 \times 10^{-4}({T} )$ & $2.1 \times 10^{-1}({T} )$ & $7.7 \times 10^{-5}({T} )$ \\
Prefix & $1.3 \times 10^{-5}({T} )$ & $1.3 \times 10^{-2}({T} )$ & $1.2 \times 10^{-3}({T} )$ & $2.2 \times 10^{-4}({T} )$ & $2.3 \times 10^{-3}({T} )$ & $1.2 \times 10^{-5}({T} )$ \\
Scratch & $1.0({T} )$ & $1.0({T} )$ & $1.0({T} )$ & $1.0({T} )$ & $1.0({T} )$ & $1.0({T} )$ \\
Zero Shot & $0.0({T} )$ & $5.6 \times 10^{-7}({T} )$ & $0.0({T} )$ & $2.0 \times 10^{-2}({T} )$ & $8.8 \times 10^{-8}({T} )$ & $8.3 \times 10^{-3}({T} )$ \\
\hline
\end{tabular}
}
\label{tab:group_performance_obstacle}
\end{table}

\section{Prefix Tuning Additional Results}
\label{sec:appendix_prefixadditional}

\begin{figure}[h!]
    \vspace{5mm}
    \centering
    \includegraphics[width=.8\textwidth]{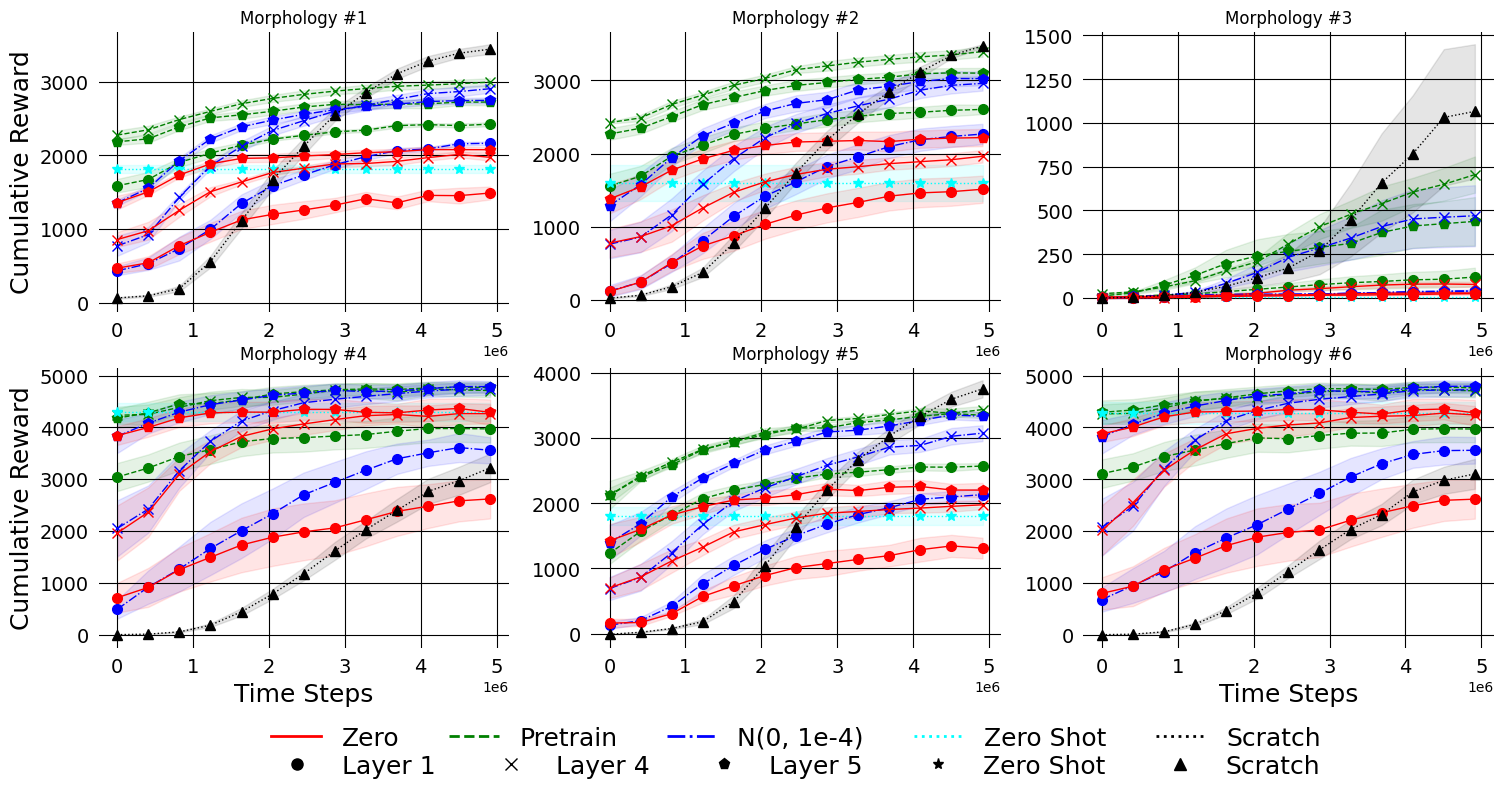}
    \caption{Choice of initialization and injection layers of prefix tuning in flat terrain.  Initial zero-shot results of E2E learning are plotted to compare affect of prefixes.}
    \label{fig:prefix-injection-layer-ablation-ft}
\end{figure}

\begin{figure}[h!]
    \vspace{5mm}
    \centering
    \includegraphics[width=.8\textwidth]{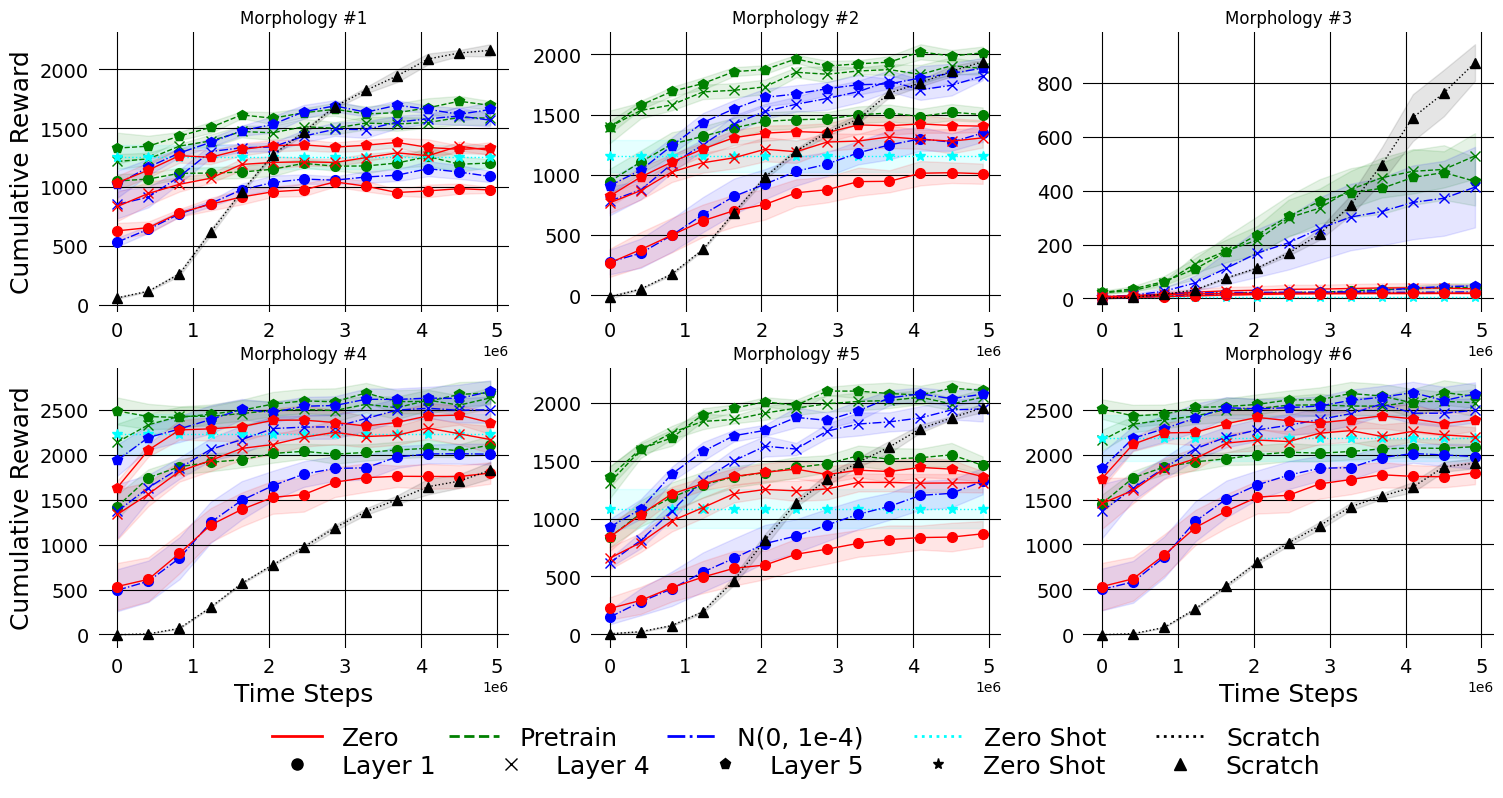}
    \caption{Choice of initialization and injection layers of prefix tuning in obstacle avoidance.  Initial zero-shot results of E2E learning are plotted to compare affect of prefixes.}
    \label{fig:prefix-injection-layer-ablation-obstacle}
\end{figure}

\begin{figure}[h!]
    \centering
    \begin{subfigure}[b]{0.48\textwidth}
        \centering
        \includegraphics[width=.8\textwidth]{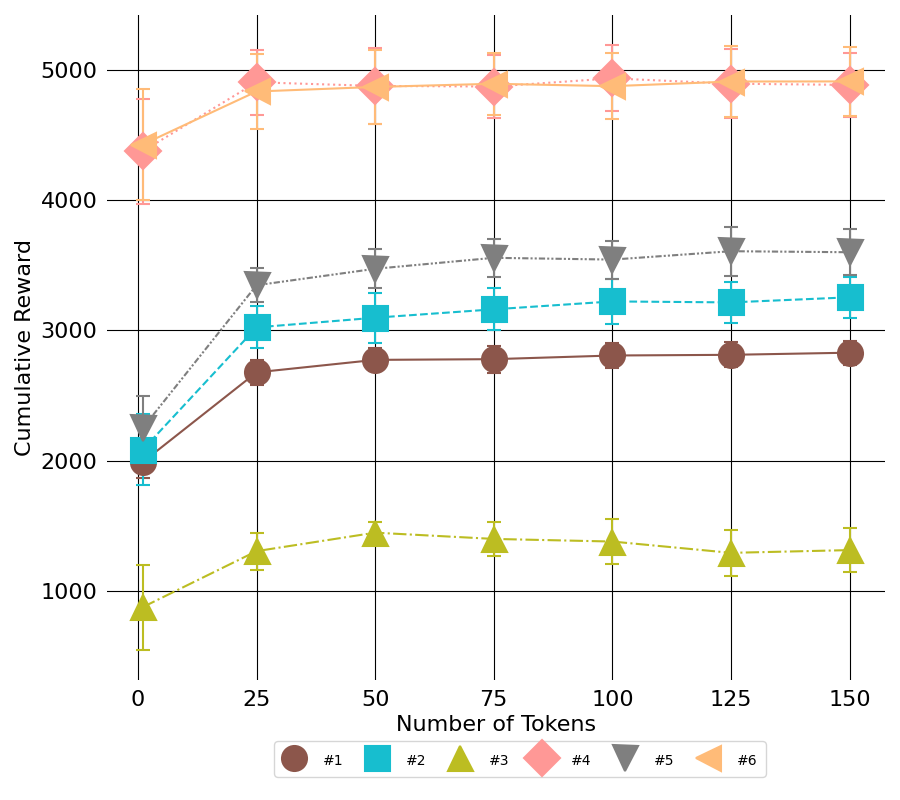}
        \caption{Number of randomly initialized prefix tokens} 
        \label{fig:ablation-n-tokens-flat}
    \end{subfigure}
    \hfill
    \begin{subfigure}[b]{0.48\textwidth}
        \centering
        \includegraphics[width=.85\textwidth]{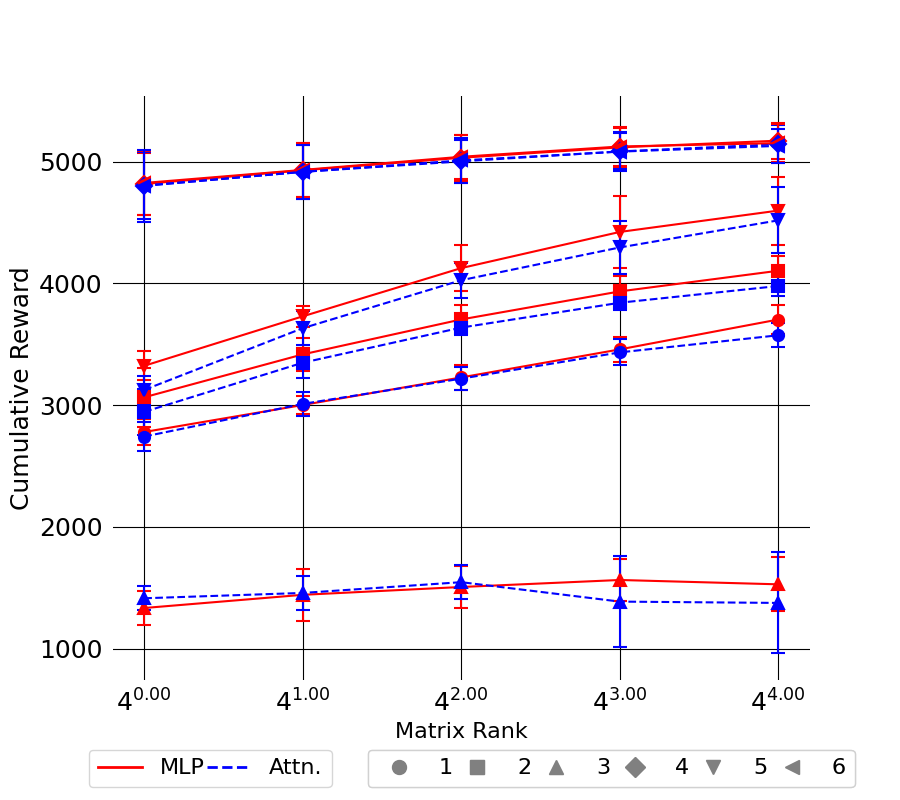}
        \caption{Lora in different layers of fifth transformer block. }
        \label{fig:ablation-lora-layer-flat}
    \end{subfigure}
    \caption{Ablation studies on number of prefix tokens and LoRA  in flat terrain task. } 
    \label{fig:combined-ablation-flat}
\end{figure}

\begin{figure}[h!]
    \centering
    \begin{subfigure}[b]{0.48\textwidth}
        \centering
        \includegraphics[width=.8\textwidth]{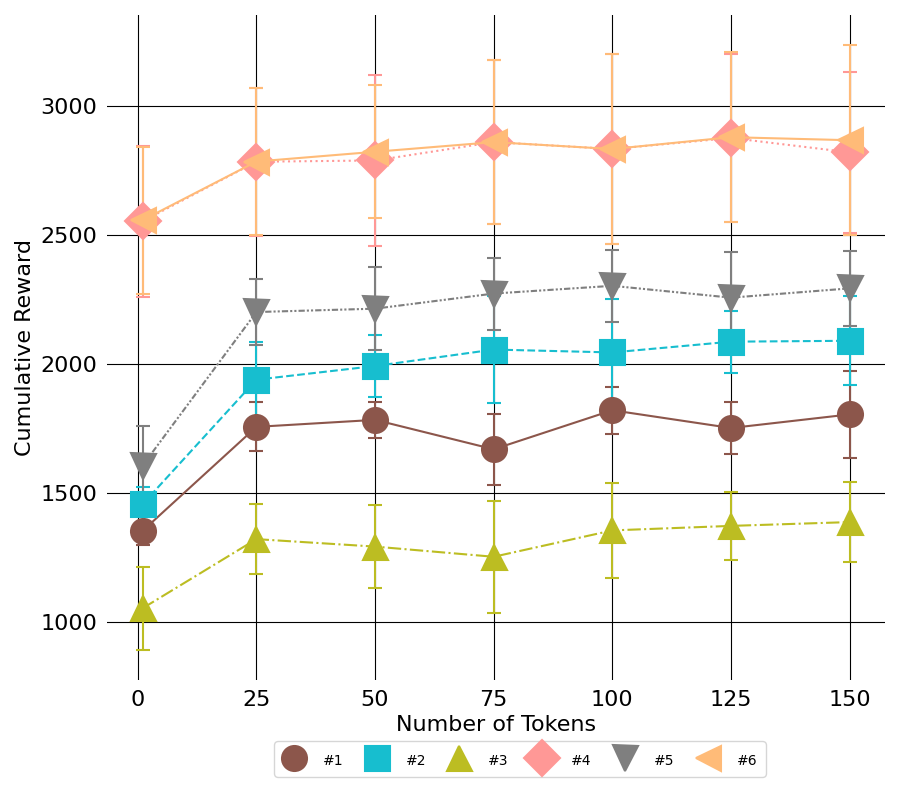}
        \caption{Number of randomly initialized prefix tokens} 
        \label{fig:ablation-n-tokens-obstacle}
    \end{subfigure}
    \hfill
    \begin{subfigure}[b]{0.48\textwidth}
        \centering
        \includegraphics[width=.85\textwidth]{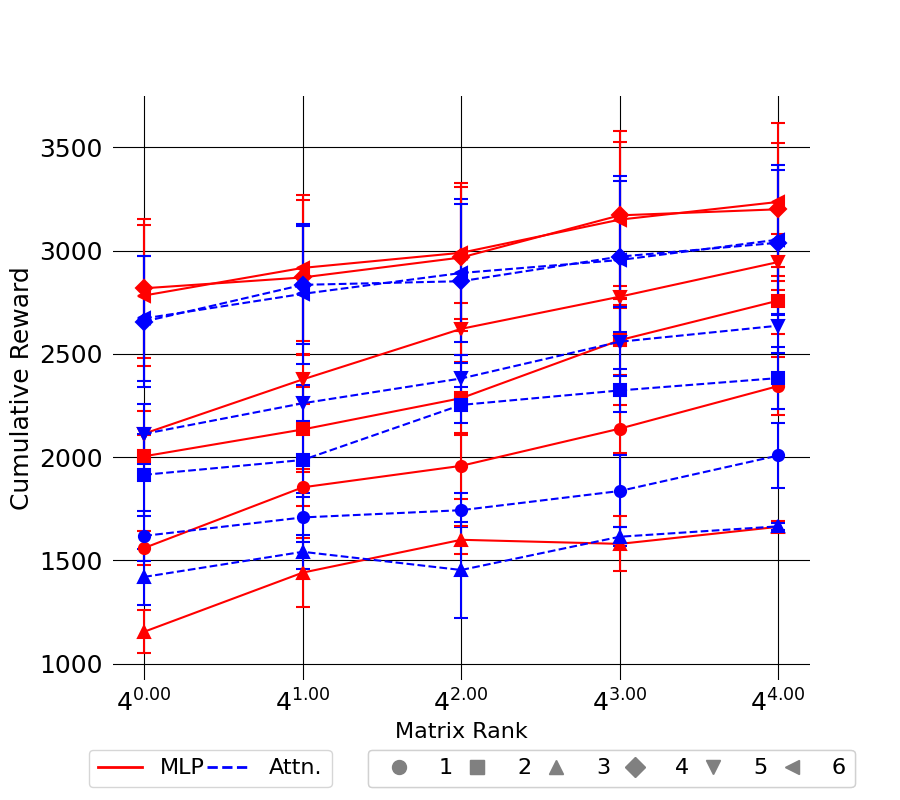}
        \caption{Lora in different layers of fifth transformer block. }
        \label{fig:ablation-lora-layer-obstacle}
    \end{subfigure}
    \caption{Ablation studies on number of prefix tokens and LoRA  in obstacle avoidance task. } 
    \label{fig:combined-ablation-obstacle}
\end{figure}

In this section, we include plots similar to those in the main paper for our prefix-tuning ablation experiments. Flat terrain results are shown in Figure~\ref{fig:prefix-injection-layer-ablation-ft} and obstacle avoidance in Figure~\ref{fig:prefix-injection-layer-ablation-obstacle}. We also show similar ablation results for LoRA and prefix tuning for flat terrain in Figure~\ref{fig:combined-ablation-flat} and obstacle avoidance in Figure~\ref{fig:combined-ablation-obstacle}. 

\end{document}